\documentclass{article}

\PassOptionsToPackage{numbers, sort&compress}{natbib}

\usepackage[preprint]{neurips_2020}

\usepackage{natbib}

\usepackage[utf8]{inputenc} % allow utf-8 input
\usepackage[T1]{fontenc}    % use 8-bit T1 fonts
\usepackage{hyperref}       % hyperlinks
\usepackage{url}            % simple URL typesetting
\usepackage{booktabs}       % professional-quality tables
\usepackage{amsfonts}       % blackboard math symbols
\usepackage{nicefrac}       % compact symbols for 1/2, etc.
\usepackage{microtype}      % microtypography

\usepackage{graphicx}
\usepackage{xcolor}
\usepackage{authblk}
\usepackage{amsmath}
\usepackage{multirow}
\usepackage{subcaption}
\usepackage{amsfonts, amstext, amsmath, amssymb, amsthm}

\usepackage{algorithm}
\usepackage{algorithmic}

\usepackage{makecell}
\usepackage{longtable}
\usepackage{adjustbox}
\usepackage[normalem]{ulem}
\useunder{\uline}{\ul}{}

\usepackage{pgfplots}
\usepackage{tikz}
\usetikzlibrary{positioning}

\usepackage[usestackEOL]{stackengine}
\usepackage{framed}
\usepackage{mdframed}
\mdfsetup{font=\small}
\usepackage[most]{tcolorbox}

\title{Making Logic Learnable With Neural Networks}

\renewcommand\footnotemark{}

\author[1, 2]{\textbf{Tobias Brudermueller}\thanks{Email: tobias.brudermueller@rwth-aachen.de}}
\author[2]{\textbf{Dennis L. Shung}}
\author[3]{\textbf{Adrian J. Stanley}}
\author[1]{\textbf{Johannes Stegmaier}}
\author[2]{\textbf{Smita Krishnaswamy}\thanks{Email: smita.krishnaswamy@yale.edu}}

\affil[1]{Institute of Imaging and Computer Vision; RWTH Aachen University; Aachen, Germany} 
\affil[2]{Yale School of Medicine; Yale University; New Haven, CT, USA}
\affil[3]{Glasgow Royal Infirmary; Glasgow, United Kingdom}

\begin{document}

\maketitle

\vspace{-6mm}
\begin{abstract}
\label{abstract}

While neural networks are good at learning unspecified functions from training samples, they cannot be directly implemented in hardware and are often not interpretable or formally verifiable. On the other hand, logic circuits are implementable, verifiable, and interpretable but are not able to learn from training data in a generalizable way. We propose a novel logic learning pipeline that combines the advantages of neural networks and logic circuits. Our pipeline first trains a neural network on a classification task, and then translates this, first to random forests, and then to AND-Inverter logic. We show that our pipeline maintains greater accuracy than naive translations to logic, and minimizes the logic such that it is more interpretable and has decreased hardware cost. We show the utility of our pipeline on a network that is trained on biomedical data. This approach could be applied to patient care to provide risk stratification and guide clinical decision-making. 
% \textcolor{red}{[TODO: Check author footnote signs]}
\end{abstract}
\vspace{-4mm}
\vspace{-1mm}
\section{Introduction}
\label{introduction}

Neural networks, equipped with the benefits of differentiable computing, have proven to be effective in learning many types of functions that translate from complex inputs to abstract outputs using training data \cite{lin1991complexity}. However, differentiable computing is not the native language of computers, since they are built using digital logic. This discrepancy makes it difficult to implement neural networks directly in hardware \cite{misra2010artificial}. Moreover, the language of logic has a distinct advantage in verifiability. When a neural network makes a decision, it is often considered a “black box”, whereas a chain of logical operations can be verified for correct operation in important conditions \cite{BRUYNOOGHE199191, 10.5555/911017, setiono2000opening}. On the other hand, there has been  little work in learning logic \cite{louis2002learning, cheang2007applying, louis2005genetic}. Most efforts in logic synthesis focus on minimizing logic that starts out specified \cite{chatterjee1995lot, brayton1984logic, lysecky2003chip, sapra2003sat, hassoun2012logic}. In this work, we seek to combine the learning process of neural networks with the advantages of logic to aid healthcare applications. Here, we bridge the gap between these two paradigms  by proposing a process for learning a  neural network and optimizing the resultant logic in a specific way that maintains accuracy. 

Our method first learns a neural network for a classification decision, and translates it first to a random forest, and then to a network of simple logic gates called an And-Inverter-Graph (AIG) \cite{abc}. We hypothesize that random forests naturally eliminate dependencies based on unseen inputs in such a way that the coverage of the logic function expands smoothly, i.e. allows us to apply the same logic decision to be applied to test vectors that are near to training vectors based on Hamming distance. Thus, we smooth the logic in this way to be more generalizable. We show that this process not only results in more efficient logic, but vastly improves the accuracy and generalizability of the logic on test data over straightforward translations to arithmetic circuits. We also show that it improves on LUT-based translations to logic that are often used in FPGAs. 

We explore logic learning and synthesis in the healthcare realm. In this realm there is a pressing need for a rational, data-driven approach to decision-making. For instance, decisions of whether to give certain treatments to patients, or predictions of which patients will require intensive care unit (ICU) care or ventilation can help hospitals plan and allocate resources that we see can become scarce under pandemic conditions. Further, black-box decisions are less trusted in healthcare even if they lead to accurate results. Therefore, being able to create a report of the decision as a logical inference has high value for verification and interpretation of the decision. For example instances that satisfy the criteria for a positive decision (positive disease diagnosis or intervention decision) can be generated using a SAT solver like MiniSAT \cite{een2003extensible}. These solutions can be verified by clinicians either using domain knowledge or historical data. 

Finally, recently there have been efforts in automated and continual predictions of medical diagnoses and interventions. For example continuous monitoring of respiratory status and blood oxygen levels can immediately identify critical care need \cite{labonchip} that are being implemented directly in hardware. In such cases the smaller and and more accurate logic we propose can improve online healthcare decisions in terms of hardware size and power consumption. While there are several methods that implement neural networks into logic \cite{misra2010artificial, wang2019deep}, as we show, direct compilations of logic can result in inaccuracy. 

We train logic networks to make healthcare decisions on two datasets. The first dataset consists of 3,012 patient observations obtained from hospitals around the world \cite{stanley2017comparison} who seek medical care for gastrointestinal bleeding (GIB) symptoms. A large proportion of GIB symptoms are not serious and most patients can be discharged. However, a small percentage of patients have serious issues that could result in death and must be hospitalized. Distinguishing between these two types of patients on the basis of the first hours of measurement is vital for hospital resource allocation. The second dataset is from the Veterans Aging Cohort Study (VACS) \cite{vacscite}  that collects detailed medical data about HIV infected veterans, when they visit the hospital to receive antiretroviral treatment. Here, we use a data set with 7,346 total observations to classify if the patient will survive the treatment for 6, 7 or 8 years.

In both cases we observe that the neural network has high accuracy, but the accuracy is diminished during the quantization and translation to arithmetic logic. Further, we often see an increase in the size of the resultant logic as measured by the number of basic AIG nodes (a standard approach in logic synthesis). We anticipate that the majority healthcare applications will continue to be implemented as software neural networks for the time being. However, we hope to provide a foundation for understanding how to translate from a neural network representation to a logic representation in a principled way that maintains or adds generalizability  and limits logic bloat, which would lead to area and power inefficiencies. 
% \newpage
\vspace{-3mm}
\section{Background}
\label{background}

Here, we focus on explaining logic-related methods that we use. Additionally, we provide some basics about Boolean logic in Section \ref{boolean_logic_background} of the appendix. Besides its written form, a Boolean function can be represented by a directed acyclic graph (DAG) in which each node models a logic gate and each edge models a gate connection. A specific implementation of a Boolean function as a DAG is the so-called \emph{And-Inverter-Graph (AIG)}, for which both AND- and Inverter-gates are part of a universal gate set. In this context, a tool that we use is \emph{ABC} \cite{abc}, which provides powerful transformations, such as redundancy removal that leads to a reduced number of nodes and levels. For instance, when AIG rewriting is performed in ABC, a minimal representation is found among all decompositions of all structural cuts in the AIG, while global logic sharing is captured using a structural hashing table \cite{abc}. In ABC the gates are also factored such that they only receive two inputs and that the nodes follow a topological order. The \emph{area or size} of an AIG is given by the number of the nodes in the graph, while the \emph{depth or delay} is given by the number of nodes on the longest path of the primary inputs and outputs. %maybe tell something about the operations it uses

A \emph{look-up table (LUT)} is a table that saves previously calculated results or information in form of array-like entries that are easily accessible. In Boolean logic, an $N$-bit LUT can encode a Boolean function with $N$ inputs by storing the corresponding truth table. Thus, in the case of $N$ bits it has $2^N$ rows, one row for each possible bit-pattern. In digital logic, a LUT is implemented with a \emph{multiplexer (MUX)}, which has select lines. Those are driven by the inputs to the Boolean function to access the value of the corresponding output that is stored in the array. Unlike multiplexers, other arithmetic circuit pieces exist to fulfill the task of adding two Boolean numbers, multiplying them or comparing them against each other. The components to realize such calculations are called \emph{adder}, \emph{multiplier}, and \emph{comparator} respectively. A method which we use in this paper, to which it can be referred as \emph{LogicNet}, is formed by a combination of look-up tables \cite{learningandmem}. Those LUTS are arranged in successive layers, similar to how it is done in neural networks. But a key difference is that the training process does not involve a backpropagation and instead rather is a memorization process. Each LUT in a layer receives inputs from only a few LUTs in the previous layer, for which the connections are chosen at random. The number of outcome columns of a single LUT depends on the number of different outputs that can be observed for the input patterns. Each entry in a row in the outcome columns counts how many times the pattern is associated with the outcome in the given data set. Hence, LogicNet is a network of concatenated look-up tables with multiple layers and serves the memorization of the information given by the data, but also includes noise.

Decision trees are a popular method in machine learning, due to their interpretability \cite{hara2016making}. They are particularly suitable for classification tasks where the feature space can be separated into distinct bins which can be shown as a tree-structure. When using multiple trees in an ensemble and letting them vote on the classification, this substantially improves the accuracy, which leads to a method known as \emph{random forests}. The growing of a random forest is usually based on random vectors that govern the growth of each tree in an ensemble \cite{randomforest_literature}.

\vspace{-3mm}
\section{Related Work}
\label{relatedwork}

\textbf{Machine Learning and Logic:} The intersection of machine learning and logic is exploited in \cite{circuitoverfitting}, which showed that circuit-based simulations serve as intrinsic method to detect overfitting of a machine learning model. Contrastingly, \cite{learningandmem} investigated the trade-off between memorization and learning with LogicNet. Other work focused on the relationship of neural networks and Boolean satisfiability. For instance, \cite{bunz2017graph, selsam2018learning} use deep neural networks to learn solving satisfiability problems as an alternative to SAT-solvers. Contrastingly, \cite{xu2018fast} and \cite{prasad2007binary} use deep learning to solve problems related to binary decision diagrams, such as variable ordering or estimation of state complexity.

\textbf{Logic Synthesis and Hardware Implementations:} There is a substantial amount of work about implementing already trained neural network architectures on Field Programmable Gate Arrays (FPGAs), such as the frameworks provided in \cite{fpgaconvnet}, which employs a synchronous dataflow model of computation, \cite{jung2007hardware} which allows real-time control of the backpropagation learning algorithm or \cite{wang2019lutnet} where the FPGAs' native LUTs are used as inference operators. In a regular setting of neural networks being compiled into hardware \cite{embeddedbnn,accbcn,fraser2017scaling,finn} the starting point are binarized neural networks \cite{hubara2016binarized}, which have binarized weights and activations during training and thus, have a different training behaviour than standard neural networks. To the best of our knowledge, there is no work to learn logic gate structures by using quantized neural network activations, and compiling them into AIGs.

\textbf{Advances in Health Care:} According to \cite{topol2019high}, the advances of deep learning directly affect the field of medicine and health care at all three levels: clinicians, health systems, and patients. Especially, for many diseases, accurate and timely clinical decision making is critically required. Lengthy and costly procedures are, however, a bottleneck at this point and can potentially be addressed by low-cost diagnostic tools that run on chips, close to where sensors produce data. Hence, there is a high demand for miniaturization and so-called \emph{lab-on-chip (LoC) technology} \cite{labonchip}. We address this issue in our work by providing a novel pipeline to translate neural networks into logic gates which can serve as further input to industrial logic synthesizers to produce on-chip machine learning designs.

\textbf{Interpretability:} In their work, \cite{murdoch2019interpretable} define \emph{interpretability} in the context of machine learning as ``the use of machine-learning models for the extraction of relevant knowledge about domain relationships contained in data." By means of this, relevance refers to knowledge and insights that are needed by an audience in a specific domain problem. For that reason, in some cases interpretable models that are less accurate than non-interpretable ones might even be preferred \cite{ribeiro2016model}. In \cite{int_techniques}, it is distinguished between \emph{intrinsic and post-hoc interpretability}. The first refers to self-explanatory models due to their structures, which for example includes decision trees and other rule-based learning methods. The latter involves deriving a second model for explaining the first one, which our work of deriving random forests, look-up tables and logic from neural networks belongs to. Our novel pipeline is also inspired by \cite{localexplanation}, where gradient-boosted decision trees that provide local explanations are combined to represent global structures of a model. They show that especially on medical data this can leverage rich summaries of both an entire model and individual features. 
\vspace{-3mm}
\section{Methods}
\label{methods}

% -------------------------------------------------------
\vspace{-2mm}
\subsection{Generalization with Don't Care Dependency Elimination}
\label{dont-care-elimination}
Traditionally, logic synthesis involves optimizing a completely specified circuit. The most basic form of specification is a truth table which lists every binary input combination and its specified output. However, in the case of "learning logic" we are in a realm where the output for every input is not specified but rather, there are holes or {\em don't cares} in the truth table. While the logic circuit representing this subset of inputs is only guaranteed to be correct on specified inputs, we hypothesize that there are ways to optimize this logic such that it does generalize, i.e. fill in holes in the truth table in an accurate way. This is what we cover in the following. 

First, we note that a logic operation called {\em dependency elimination} offers a K-nearest-neighbor-like interpolation based on cube expansion. Building on this observation, we use random forests as an intermediate step of our pipeline. Second, factored forms like multi-levels of LUTs can be seen as a way of abstraction and "clever" memorization as a learning process. Given this motivation, we use LogicNets \cite{learningandmem} as alternative intermediate step of our pipeline and as one of multiple baselines. 

The most direct way to obtain a logic circuit is as a \emph{sum-of-products (SOP)} which is a two-level logic obtainable directly from the truth-table. Given an $n$-input logic function $f(x_1, \ldots , x_n)$ the circuit consists of an AND (product or cube) gate $AND_i$ for each vector $\vec{x_i}$ with $n$ entries such that $f(\vec{x_i})$ is 1, and there is a single $OR$ gate that takes each of the AND gates as input. Thus, for $k$ observations $\vec{x_i}$ it follows: $f(\vec{x_i}) = f(x_1, \ldots, x_n) = OR(AND_1, \ldots, AND_k)$. Note that this basic form of logic is exactly memorization in that each training vector that produces a $1$ is encoded directly as an AND gate, thus this function does not generalize outside the training data. This circuit has high complexity in that an $n$-input AND gate requires a tree of $2$-input AND gates for hardware implementation. Thus, an $n$-input AND gate will require roughly $n$ 2-input AND gates. 

The primary goal of logic synthesis involves reducing the number of basic gates for building the circuit. One basic operation in logic synthesis is dependency elimination. In an SOP setting an example of dependency elimination is cube expansion, e.g. $ab+a{\bar{b}}= a(b+{\bar{b}}) = a$. Here, the input $b$ is eliminated due to lack of dependency on its value. 
% In the case of unseen inputs, this is termed {\em don't care-based dependency elimination} in the following and can lead to generalization.

Suppose that $\mathcal{B} = \{0, 1\}$ is a Boolean space. Then, for an n-input logic function \mbox{$f(x_1, \ldots, x_n): \mathcal{B}^n \to \mathcal{B}^1$}, {\em single-variable dependency elimination} is defined as applying Equation \ref{dontcare-example} if both $f(\vec{x_i})=1$ and $f(\vec{x_i})=0$ are not seen in combination with the other literals: 
% \vspace{-1mm}
\begin{equation}
\label{dontcare-example}
\displaystyle
 f(x_1, \ldots, x_n) = f(x_1, \ldots,  x_{i-1}, 0, x_{i+1}, \ldots, x_n) = f(x_1, \ldots, x_{i-1}, 1, x_{i+1}, \ldots, x_n)
 \end{equation}
% \vspace{-1mm}

\textbf{Proposition:} Eliminating $n$ variables in a logic function $f$ results in a $2^n$ increase in logic coverage.

\textbf{Proof:} Each variable eliminated doubles the size of the function, because now the function applies to all inputs as before and their corresponding inputs with the eliminated variable negated. So inductively, if we eliminate one variable at a time, we double the coverage each time.  Note that the coverage increases inputs that are close in Hamming distance to those already covered by the function $f$ and therefore, this is a smooth expansion. 

Suppose we have a set of training samples $\mathcal{X} = \{\vec{x_1}, \ldots, \vec{x_m} \}$, where $\vec{x_i} = [x_{i,1}, x_{i,2},  \ldots, x_{i,n}]^T$. Suppose the corresponding test data $\mathcal{T} = \{\vec{t_1}, \ldots, \vec{t_q} \}$ is distributed around the training data and generated by a bit flip on any bit of the training data. Then start with two SOP circuits: $C_1= \sum P_i$ where $P_i=\Pi \vec{x_i}$ such that $f(\vec{x_i})=1$ and $C_2= \sum N_i$ with, $N_j=\vec{x_j} $, $j \neq i$ such that $f(\vec{x_j})=0$. Then derive the response to a test input $\vec{t_i}$ as follows:
\begin{equation}
\label{test_assignment_example}
\displaystyle{
f(\vec{t_i}) = 
 \begin{cases}
    1 & \textnormal{if } C_1(\vec{t_i})==1 \textnormal{ and } C_2(\vec{t_i})==0\\
    0 & \textnormal{if } C_1(\vec{t_i})==0 \textnormal{ and } C_2(\vec{t_i})==1 \\
    \textnormal{randomly pick 0 or 1} & \textnormal{otherwise}
\end{cases}
}
\end{equation}

Now suppose we have a new circuit $C'_1= \sum P'_i$ that randomly eliminates one input $P_i$ in $C_1$ such that the test data at a 1-bit Hamming distance from the training data receives the same answer. Suppose to do the same for $C_2$ to create $C'_2$. This also means that each of $P_i$ and $N_i$ cover two nearby vectors rather than one: one from the training set and one from the test set. Thus, the close test set gets assigned to the same value as the training set. We term this {\em don't care-based dependency elimination} that can lead to generalization, as we eliminated dependencies with respect to unseen inputs. 

To connect this to a random forest, we note that random forests consist of decision trees. Decision trees are effectively logical AND statements (products). For instance, if the branch $v_1=1, v_2=1, v_3=1$ leads to a decision of $1$, then this is an AND statement $B=v_1v_2v_3$, and thus the random forest can be thought of as SOP with products corresponding to each branch. Since decision trees do not decide (i.e. branch) on variables that do not affect the decision, corresponding variables are eliminated from the products leading to generalization on unseen inputs based on logic smoothness. 

% -------------------------------------------------------
\vspace{-2mm}
\subsection{Generalization with Factorization}
\label{factorization_generalization}

Another logic operation that has been shown to generalize is {\em factorization}, which involves adding depth in order to pull out a common term in logic. An example of factorizing is $abc+abd = ab(c+d)$, where the factor $ab$ is pulled out of both terms of the sum. Again cube-expanding this expression would allow an inductive generalization argument where based on $ab$ being the decisive factor in some examples it is generalized to others. The work of \cite{learningandmem} shows that factorized LUTs also generalize over simple LUTs. This motivates to include translations from a neural network to a high-depth LUT network - {\em LogicNet} - as one of our baselines. 
% It is hypothesized that both pipelines, the one including LogicNet and the one building on random forests, lead to lower complexity as well as improved accuracy over the direct translation. 

% -------------------------------------------------------
\vspace{-2mm}
\subsection{Logic Verifiability}
\label{logic_interpretation_methods}
Logic has additional advantages, particularly in the health care realm. First, if a diagnosis or treatment decision is decided by a logic circuit, then a {\em report} can be created such that the decision can be explained by a logic diagram. Second, using logic allows for the use of SAT solvers, whose speed and performance have been the key factors for success in modern logic synthesis. For a Boolean function $f(\vec{x})$ a SAT solver, such as MiniSAT in ABC \cite{abc,een2003extensible}, can find an input vector $\vec{y}$ such that $f(\vec{y})=1$. The input vector $\vec{y}$, thus, satisfies the logic result. Such vectors can be generated en masse to test features of the system and potentially add features or change the data set design if the decision is not medically valid. Further, in any such input vector one can analyze the controlling inputs, meaning those whose bit flip causes the decision to change. This helps in determining the critical and potentially causative factors in the logic.

\textbf{Example:} For the Gastrointestinal bleeding data that we consider, the elevation of blood urea nitrogen above 18.2 in a clinical setting suggests volume loss or reabsorption of blood in the gastrointestinal tract. Systolic blood pressure below 100 mmHg in the same clinical setting likewise suggests volume loss. Hemoglobin less than 10 g/dL in females reflects an abnormally low blood count even taking into account the possibility of regular bloodloss from menses. Each of these statements describes a condition that reflects an abnormal amount of bloodloss, which requires medical attention. This statement can be ported into logic as follows: \emph{(Blood Urea Nitrogen > 18.2 mg/dL) OR (Systolic Blood Pressure <100 mmHg) OR (Hemoglobin < 10 g/dL AND NOT Male)}.
This is in conjunctive normal form in that it is a conjunction of single disjunctions. A SAT solver could satisfy this by generating solutions which involve setting any of the 3 variables to true. 

% -------------------------------------------------------
\vspace{-2mm}
\subsection{The Framework}
\label{framework}

Our framework starts with a trained real-valued neural network and then trains random forests on the nodes' quantized activations before a translation to logic. While we provide a detailed description of the logic creation process in Section \ref{logic_creation_description} of the appendix, Figure \ref{decision_tree_figure} shows an example of the logic of a single decision tree. Our pipeline consists of the following steps:
\begin{enumerate}
    \item A neural network is trained on a classification task with real-valued inputs ${\displaystyle\vec{x_i}}$ and labels ${\displaystyle t_i}$ of a training data set ${\displaystyle \mathcal{X}_t}$. Each layer ${\displaystyle l}$ with ${\displaystyle N_l}$ nodes has a set of real valued activations ${\displaystyle \mathcal{A} =\{a_{l;1}, a_{l;2} \ldots a_{l;{N_l}}\}}$ and weights ${\displaystyle \mathcal{W} = \{\vec{w_{l;1}}, \vec{w_{l;2}} \ldots \vec{w_{l;{N_l}}}\} }$. \vspace{-0.5mm}
    
    \item At each node ${\displaystyle n}$ of each layer ${\displaystyle l}$ each real valued activation ${\displaystyle a_{l;n}}$ and weight vector ${\displaystyle \vec{w_{l;n}}}$ is quantized according to a quantization scheme into ${\displaystyle m}$ bits to gain ${\displaystyle a_{l;n}^q}$ and ${\displaystyle \vec{w_{l;n}^q}} $ accordingly (see Section \ref{quantization}).\vspace{-0.5mm} 
    
    \item Let ${\displaystyle l=0}$ be the input layer and ${\displaystyle l=L}$ be the output layer of the neural network. For each node ${\displaystyle 0{\leq}n{<}{N_l}}$ in layer ${\displaystyle 0{<}l{<}L}$, a random forest ${\displaystyle RF_{l;n}}$ is trained on a data set ${\displaystyle Z_{l;n}}$ that consists of quantized activations ${\displaystyle \{{a_{[l-1];k}^q} \mid {\forall} k{\in}[0,{N_{l-1}]}\}}$ from the previous layer as features and the corresponding quantized activation ${\displaystyle a_{l;n}^q}$ from node ${\displaystyle l;n}$ as label.\vspace{-0.5mm}
    
    \item Now, each ${\displaystyle RF_{l;n}}$ is translated into equivalent logic ${\displaystyle RF_{l;n}^{logic}}$ (see Section \ref{rf_translation}). It is referred to this as \emph{module} or interchangeably as \emph{block}.\vspace{-0.5mm}
    
    \item Each logic module is cascaded together to reform the entire neural network structure, which yields the full logic ${\displaystyle Logic_{RF}}$.\vspace{-0.5mm} 
    
    \item Lastly, an And-Inverter-Graph ${\displaystyle AIG_{RF}}$ is created from ${\displaystyle Logic_{RF}}$ using the ABC tool \cite{abc}. It can be used to measure size and complexity and for simulation purposes and thus, for gaining insights on the performance and generalization capability of the circuit.\vspace{-0.5mm}
\end{enumerate}

\begin{figure*}[ht!]
\centering 
\includegraphics[width=0.9\textwidth]{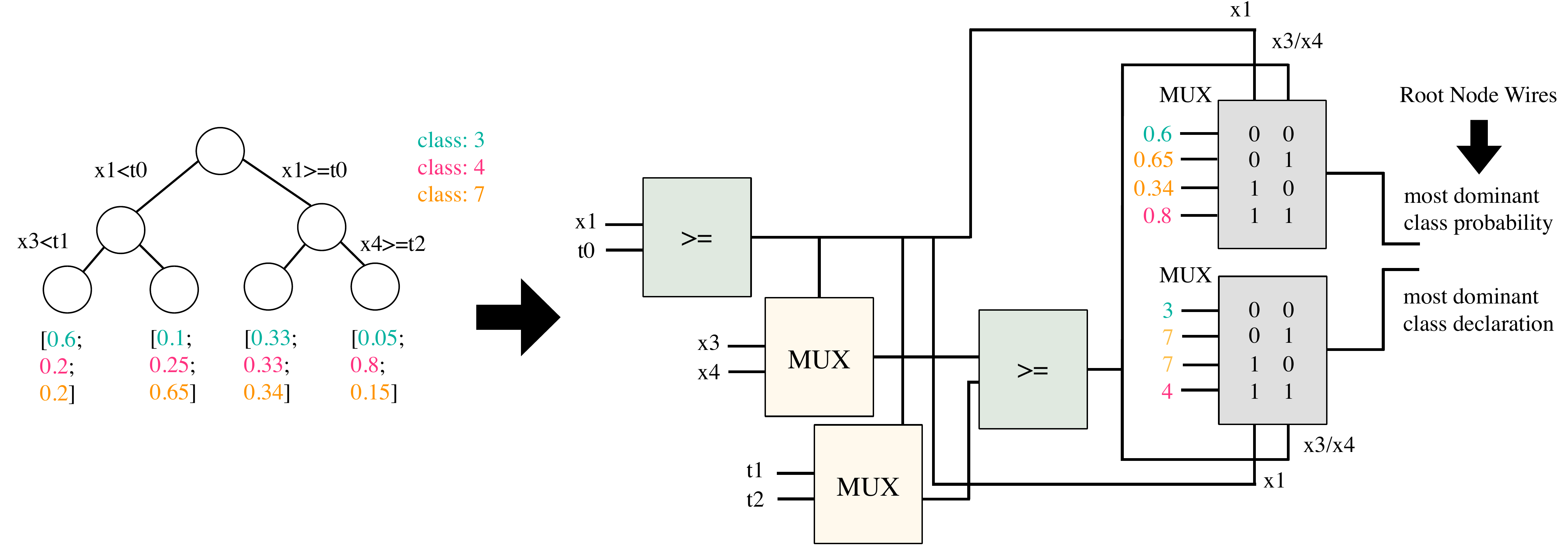}
\caption[]{Example of a decision tree logic.}    
\label{decision_tree_figure}
\end{figure*}

% -------------------------------------------------------
\vspace{-2mm}
\subsubsection{Quantization Scheme}
\label{quantization}
The neural network's weights, activations, and the data, are typically floating point values that have a limited precision of 32 or 64 bits. Here, it is desired to derive binary strings from these floats with a lower precision of $m$ bits, while maintaining information from before and after the decimal point and about the sign, which needs to be done by binning and clipping in combination with a two's complement binarization. Therefore, the number of total bits for the quantization is defined as $m = k+i$ with $k$ being the number of bits that represent the information before the decimal point (\emph{integer bits}) and $i$ for the positions after the decimal point (\emph{fractional bits}). The quantization scheme is used consistently for every value that needs to be quantized. The pseudo-code can be found in Algorithm \ref{quantization_algo}.

% \vspace{-5mm}
\begin{algorithm}[ht]
   \caption{Quantization Scheme}
   \label{quantization_algo}
\begin{algorithmic}
   \STATE {\bfseries Input:} float value $x$, number of total bits $m$, number of fractional bits $i$
   \STATE {\bfseries Conversion to Integer-Representation:} 
   \STATE $x_{int} = int(1<<i)*x$
   \STATE $largest\_signed\_int = (1<<(m-1))-1$
   \STATE $x_{int} = min(largest\_signed\_int, x_{int})$
   \STATE $smallest\_signed\_int = -(1<<(m-1))$
   \STATE $x_{int} = max(smallest\_signed\_int, x_{int})$
   \STATE $largest\_unsigned\_int = (1<<m)-1$
   \STATE $x_{int} = x_{int} \& largest\_unsigned\_int$
   \STATE {\bfseries Conversion to Binary String:}
   \STATE return $format(x_{int}, 'b').zfill(m)$
\end{algorithmic}
\end{algorithm}
% \vspace{-5mm}
\vspace{-2mm}
\section{Evaluation}
\label{evaluation}
% \vspace{-2mm}

% -------------------------------------------------------
\vspace{-2mm}
\subsection{Gastrointestinal Bleeding (GIB) Data Set}
The evaluation of our proposed methods is performed on two different biomedical data sets of classification problems. The first one, namely the GIB data set collects information about patients presenting with \emph{gastrointestinal bleeding (GIB)} at multiple hospitals around the globe \cite{stanley2017comparison}. The data set contains 3,012 observations in total with 27 numerical or categorial features for each observation. The features cover general information about the patient, medical values and even include previous medication. As outcome variable for a binary classification task we use the doctors' decisions for a need of immediate hospital-based intervention. The data set is generally balanced in terms of the two classes and without missing values. For the test data, a perfectly balanced split of 20\% of the overall data is used, meaning that random guessing would yield 50.00\% accuracy. Our method achieves 83.28\% accuracy, where an arithmetic circuit only achieves 70.36\%.

% -------------------------------------------------------
\vspace{-2mm}
\subsection{Veterans Aging Cohort Study (VACS) Data Set}
\vspace{-2mm}

The Veterans Aging Cohort Study (VACS) \cite{vacscite} is a still ongoing study that collects detailed medical data about veterans with HIV infections, when they visit the hospital to receive antiretroviral treatment. Here, a subset of this data from 1999 to 2015 is used, such that it contains 10 numerical features (medical values) of dead patients that had 40-50 visits. The data set is originally formed to calculate an index that estimates the risk of mortality \cite{tate2019albumin}. Here, the data set with 7,346 total observations is used to classify if the underlying patient with an HIV infection and corresponding treatment survived 6, 7 or 8 years. We use the same splits as for the GIB data set and a perfectly balanced test data set. Thus, random guessing or a constant output would yield 33.33\% accuracy. In this setting, our method achieves 59.27\% accuracy, where an arithmetic circuit only achieves 45.20\%.

% -------------------------------------------------------
\vspace{-2mm}
\subsection{Baselines for Comparison}
\label{baselines_description}
\vspace{-2mm}

In Section \ref{methods} we motivated why we use random forests as intermediate steps before a translation to logic and introduced the framework's steps. 

In order to evaluate this pipeline, we set up two alternative baselines for comparison. 

\begin{itemize}
\item The first baseline is a "straightforward" translation of the neural network into an {\em arithmetic circuit}, where weights are multiplied with activations using multipliers and the results summed with adders. These adders and multipliers are then decomposed to form the AIG. Thus, there are no intermediary steps in this pipeline.  

\item The second baseline is inspired by FPGA architectures. Thus, instead of training random forests on the quantized activations, we train networks of look-up tables using a setup called {\em LogicNet} \cite{learningandmem}. We hypothesize that these LogicNets may also eliminate dependencies in certain cases after factorization and may be a stronger baseline than the arithmetic circuit \ref{factorization_generalization}. 

\end{itemize}

\vspace{-2mm}
\subsection{Quantization Schemes}
\vspace{-2mm}

An important step in all 3 pipelines is the initial quantization of the neural network before subsequent steps (of learning arithmetic circuits, LogicNet modules, or random forests). Both the logic complexity and the accuracy are directly linked to the number of total bits in the quantization scheme. Traditional logic holds that the more bits used, the better accuracy overall, but our results show that this is not always the case due to better optimizations that are able to be performed on simpler quantization schemes. Therefore, we choose from a set of "small" quantization schemes $\{(6, 4), (7, 3), (8, 6), (10, 6)\}$ and a set of "big" quantization schemes $\{(12, 8), (16, 10), (18, 10), (32, 16)\}$ based on the arithmetic circuit that has the best performance. This also has the added effect of strengthening the baseline for comparison purposes. We report comparison results on a "small" and "large" quantization scheme for each arithmetic circuit baseline. 

\vspace{-2mm}
\subsection{Neural Network Architectures}
\vspace{-2mm}

While larger neural networks generally learn better, smaller networks can result in smaller logic, that is easier to verify using SAT solvers, and easier to interpret. Therefore we test our pipelines using smaller node sizes and larger node sizes. Our neural networks are fully-connected networks with ReLU activation functions on hidden layers and a softmax-function on the last layer in combination with cross-entropy loss. Different architectures, optimizers and normalization modes were used to train well-performing real-valued neural networks. Two models per data set were chosen as input to the pipeline (see Table \ref{real_nn_baselines_refined}).

\vspace{-4mm}
\begin{table}[ht!]
\caption{\small Overview of the \textbf{real-valued neural networks} that serve as input to the logic pipeline.} \label{real_nn_baselines_refined}
\centering
\begin{adjustbox}{width=0.85\textwidth}
\begin{tabular}{ccccc}
\toprule
\makecell{Reference Key} & \makecell{Hidden Layer Nodes} & \makecell{Optimizer} & \makecell{Normalization Mode} &  \makecell{Accuracy (in \%)} \\
\midrule
GIB Model 1 		&	5 - 4 - 3		&	Adam 	&           Robust-Scaling 				&  	83.28 \\
GIB Model 2 		&	12 - 10 		&	Adam 	&           Robust-Scaling 			&  	85.93 \\
\midrule
VACS Model 1		&	10 - 5	 	&	Adam 	&           Z-Score Normalization 			& 	63.20\\
VACS Model 2 		&	40 - 20 		&	RAdam	&           Robust-Scaling 			& 	75.58 \\
\bottomrule
\end{tabular}
\end{adjustbox}
\end{table}
\vspace{-2mm}

\newpage %only for arxiv version
 For LogicNet, we used the following hyper-parameters to form different configurations: \emph{depth (number of layers):} $\{2, 3, 4 \}$; \emph{width (number of LUTs in each layer):} $\{50, 100, 200, 500, 1000 \}$; \emph{LUT-size (number of inputs to each LUT):} $\{1, 2, 3 \}$. Analogously, for random forest, combinations of the following hyper-parameters were used: \emph{maximal tree depth:} $\{5, 10, 15, 20 \}$; \emph{number of estimators:} $\{2, 3, 4 \}$; \emph{bitwise training:} $\{Yes, No\}$. Note that the "bitwise training"-parameter means that we introduced the option to train the random forests bit by bit on the quantized activations or on the full bit-string. Besides the training process itself, this also impacts the logic creation. We provide more details about this in the appendix.

% -------------------------------------------------------
\vspace{-2mm}
\subsection{Results}
\label{results}
\vspace{-2mm}
Table \ref{combined_table} shows the results of the top-1 model of each configuration. It can be observed that our  approach based on random forests outperforms the arithmetic circuits not only in 6 of 8 cases in terms of accuracy, but also most of the times in logic complexity. We also note that the LogicNet approach improves over the arithmetic circuit in some cases, but on the VACS datasets it seems to degrade essentially to random guessing. Therefore it seems to have more drastic failure modes. Curiously,  there is one case in which the neural network accuracy  of 63.20\% is exactly preserved in the VACS cohort case by the arithmetic circuit. Since the VACS cohort is relatively small in terms of the number of features for each data point, it may be possible that an arithmetic circuit learns the full function accurately. However, we expect such cases to be rare. 

In Section \ref{interpretable_report} of the appendix we additionally provide an example of an interpretable report that can be derived from the logic for the purpose of system verification. There, we also deliver some run-times of MiniSAT \cite{een2003extensible} to proof that such a verification process is possible in a feasible amount of time. For a large logic setting of 912,589 AIG nodes MiniSAT allocates a maximum of approximately six hours. In most cases the run-time is in a magnitude of a few minutes to hours.

\vspace{-4mm}
\begin{table}[ht!]
\caption{\small Results of the \textbf{top-1 models} for each technique of logic creation, using the neural networks and data use-cases from Table \ref{real_nn_baselines_refined} and two quantization schemes each. Our method's values in each performance measure category are marked in bold if they outperform the ones of the arithmetic circuit. Note that a value of zero for the AIG nodes and levels means that the model has learned nothing and that the logic only delivers a constant classification output.}
\label{combined_table}
\centering
\begin{adjustbox}{width=1.0\textwidth,center}
\begin{tabular}{@{}c|c|c|ccc|ccc|ccc@{}}
\toprule
\multirow{2}{*}{\begin{tabular}[c]{@{}c@{}}Reference\\ Key\end{tabular}} &
  \multirow{2}{*}{\begin{tabular}[c]{@{}c@{}}Real-Valued\\ Network\\ Acc. (in \%)\end{tabular}} &
  \multirow{2}{*}{\begin{tabular}[c]{@{}c@{}}Quant. \\ Scheme\end{tabular}} &
  \multicolumn{3}{c|}{{\ul Arithmetic Circuit}} &
  \multicolumn{3}{c|}{{\ul LogicNet}} &
  \multicolumn{3}{c}{{\ul \textbf{NNET to Random Forest}}} \\
 &
  &
  &
  \begin{tabular}[c]{@{}c@{}}AIG\\ Nodes\end{tabular} &
  \begin{tabular}[c]{@{}c@{}}AIG\\ Levels\end{tabular} &
  \begin{tabular}[c]{@{}c@{}}Acc.\\ (in \%)\end{tabular} &
  \begin{tabular}[c]{@{}c@{}}AIG\\ Nodes\end{tabular} &
  \begin{tabular}[c]{@{}c@{}}AIG\\ Levels\end{tabular} &
  \begin{tabular}[c]{@{}c@{}}Acc.\\ (in \%)\end{tabular} &
  \begin{tabular}[c]{@{}c@{}}AIG\\ Nodes\end{tabular} &
  \begin{tabular}[c]{@{}c@{}}AIG\\ Levels\end{tabular} &
  \begin{tabular}[c]{@{}c@{}}Acc.\\ (in \%)\end{tabular} \\ \midrule
\multirow{2}{*}{GIB Model 1}  & \multirow{2}{*}{83.28} & (7, 3)   & 14,314  & 231  & 70.36   & 2,508  & 106  & 66.56  & 50,684 & 245 & \textbf{83.28} \\
                            &                    & (18, 10) & 123,389 & 451 & 83.28 &  25,536 & 198 & 65.56 & \textbf{114,011} & \textbf{357} & 77,65 \\ \midrule
\multirow{2}{*}{GIB Model 2}    & \multirow{2}{*}{85.93} & (6, 4)   & 38,414 & 240 & 52.48 & 13,501 & 131 & 64.40 & 425,617 & \textbf{213} & 56.29 \\
                            &                    & (12, 8)  & 150,427 & 338 & 45.20 & 41,867 & 165 & 50.50 & 214,009 & \textbf{300} & \textbf{59.27} \\ \midrule
\multirow{2}{*}{VACS Model 1} & \multirow{2}{*}{63.20} & (7, 3)   & 17,807 & 183 & 35.51 & 7,473 & 89 & 33.40 & 95,176 & \textbf{117} & \textbf{49.80} \\
                            &                    & (32, 16) & 338,757 & 549 & 63.20 & 186,572 & 274 & 33.40 & \textbf{254,579} & \textbf{354} & 38.91 \\ \midrule
\multirow{2}{*}{VACS Model 2}   & \multirow{2}{*}{75.58} & (6, 4)   & 106,572 & 353 & 34.97  & 0 & 0 & 33.33 & \textbf{83,697} & \textbf{146} & \textbf{37.69} \\
                            &                    & (32, 16) & 2,173,117 & 726 & 34.97 & 389,577 & 264 & 33.47 & \textbf{2,097,694} & \textbf{373} & \textbf{39.39} \\ \bottomrule
\end{tabular}
\end{adjustbox}
\end{table}

We note that AIGs themselves are minimized for logic redundancy by using a hash table to look up small patches of logic \cite{mishchenko2006dag}. However, though these operations are applied to all three models to derive the final AIG, they do not compensate accuracy and reach the same node or AIG level count. We believe that this is because the random forest models begin with a better global logic architecture, which local optimizations offered by AIGs can not correct. 

Based on our results here, we recommend the usage of the derived methods as follows: Begin with small real-valued neural networks and small quantization schemes, to derive an arithmetic circuit. When the arithmetic circuit performs poorly, our pipeline helps to increase the chance of finding a well-generalizing logic circuit that is smaller in terms of size and area and better in terms of accuracy. When the arithmetic circuit already performs well, thus forms a strong baseline, the logic learning pipeline can help finding a less bloated logic that achieves similar results.

\newpage
\vspace{-2mm}
\section{Conclusion}
\label{conclusion}
\vspace{-2mm}

Here, we presented a novel framework for translating trained neural networks first into random forests and then into logic gate representations. The motivation was to combine the learnability of neural networks with the verifiability and implementability of logic. We noted that, in general, translation of a neural network into arithmetic logic incurs a loss of accuracy. We found that the inclusion of training random forests for each activation before the translation to logic increases accuracy and decreases hardware complexity. We showed that the intermediate optimizations are a form of dependency elimination on the logic and thus, allow for exploiting logic smoothness, and improves generalization.  Thus, we provide evidence that particular logic operations themselves can generalize to unseen training examples. Future work remains in investigating the properties of our intermediate operations and different quantization schemes.

% \clearpage

\section*{Broader Impact}
\label{broader_impact}

Our work has broad applicability in many fields, and especially health care, where there is an advantage to performing logically verifiable decisions. Additionally, with our work we hope to deliver a new motivation for the machine learning community to further explore the intersection of neural networks and logic circuits. We believe that there is a high potential in translating machine learning models to logic circuits for several reasons: 

\begin{enumerate}
    \item Logic is verifiable. 
    \item Logic can be more interpretable. 
    \item Logic is implementable in hardware. Thus, if logic was learnable, machine learning models could also more easily be run on chips. 
    \item Logic could in the long-run maybe be seen as an intermediate step of rendering one machine learning model from another. This could lead to totally new concepts when it comes to intersecting multiple domains. Hence, one of the next steps could be to again derive a trainable machine learning model from logic.
\end{enumerate}

We do not believe anyone would be be put at disadvantage from this research. Nonetheless, we agree that the consequences of failure of the system would be immense, when being applied to tasks like classifying if a patient needs hospitalization. However, we do not claim that our system is free of faults at this point of research. Instead we provided a novel setup that needs to be further explored, but that can put research forth in the intersection of machine learning and logic circuits, since we motivated its exploration in a previously unseen way. We hope to trigger new ideas in the community which can have a high impact on how the aspect of interpretability in black box models is thought.

\vspace{-1mm}
% \begin{ack}
\section*{Acknowledgements}
\label{acknowledgements}
We thank Sat Chatterjee (Google AI, Mountain View, CA, USA), Alan Mishchenko (Department of EECS, University of California, Berkeley, CA, USA), and Claudionor N. Coelho (Google AI, Mountain View, CA, USA) for the discussion and the collaboration on the technical implementations. We thank the following people for the data collection: Dennis L. Shung \emph{(Yale School of Medicine, New Haven, CT, USA)}, Loren Laine \emph{(Yale School of Medicine, New Haven, CT, USA)}, Adrian J. Stanley \emph{(Glasgow Royal Infirmary, Glasgow, United Kingdom)}, Stig B. Laursen \emph{(Odense University Hospital, Odense, Denmark)}, Harry R. Dalton \emph{(Royal Cornwall Hospital, Cornwall, United Kingdom)}, Jeffrey Ngu \emph{(Christchurch Hospital, Christchurch, New Zealand)} and Michael Schultz \emph{(Dunedin Hospital, Dunedin, New Zealand)}. 

% \textcolor{red}{[TODO: Disclosure of Funding]}
% \end{ack}

\clearpage
\bibliographystyle{unsrtnat}
\bibliography{references}

\clearpage
\clearpage
\section{Appendix}
\label{appendix}

% -------------------------------------------------------
\subsection{Background: Boolean Logic}
\label{boolean_logic_background}
A \emph{Boolean equation} is a mathematical expression that only uses binary variables. \emph{Logic gates} are simple digital circuits that take a number of binary inputs to produce a binary output, i.e. they perform operations on binary variables and implement Boolean equations. An \emph{AND-Gate} for example outputs $1$ when all inputs are $1$, while an \emph{OR-Gate} outputs $1$ when any input is $1$. An \emph{Inverter-Gate} negates the input to form its \emph{complement}. A \emph{truth table} of a gate can store information about which input patterns causes which output. A concatenation of multiple gates can be used to implement a \emph{logic function}. Any Boolean variable or its complement in a Boolean equation is called a \emph{literal}. The AND of one or more literals is known to be a \emph{product} or \emph{cube}. Similarly, the OR of literals is called a \emph{sum}. In this context a \emph{sum-of-products} refers to multiple AND-terms being connected by ORs and vice versa for a \emph{product-of-sums}. Both can be transferred into each other. For a given propositional Boolean formula, the \emph{Boolean satisfiability problem (SAT)} asks whether there exists a satisfying set of variable assignments that makes the output be $1$. There are multiple \emph{SAT solvers} publicly available to fulfill this task that is linear in the number of variables \cite{een2003extensible, moskewicz2001chaff, goldberg2007berkmin}.

When using $b$ to encode $2^N$ different values, it needs to be distinguished between using the full range to encode \emph{unsigned binary numbers}, i.e. only positive values, or splitting it up into two parts that represent negative and positive numbers, hence, encoding \emph{signed binary numbers}. For the latter, the so-called \emph{two complement} representation can be used where zero is written as all zeros $00 \cdots 000_2$ in the binary representation. The most positive number is given by $2^{N-1}-1 = 01 \cdots 111_2$ and thus, has ones everywhere expect for the most significant bit. The most negative number is defined as $-2^{N-1} = 10 \cdots 000_2$ and hence, has a $1$ at the most significant bit and zeros elsewhere. Positive numbers in general have a $0$ at the most significant bit and negative numbers a $1$. Therefore, the most significant bit is in this context also called the \emph{sign bit}. However, the remaining bits need to be interpreted differently to get the magnitude of the value, which is done in a process called the \emph{two's complement}. The process is given by inverting all bits and then adding a $1$ to the least significant bit. This then turns the signed representation back into an unsigned representation and thus, yields only the magnitude of the positive or negative binary number.

% -------------------------------------------------------
\subsection{Methods: Logic Creation}
\label{logic_creation_description}
\begin{figure*}[ht]
\vskip 0.2in
\begin{center}
\centerline{\includegraphics[width=0.8\textwidth]{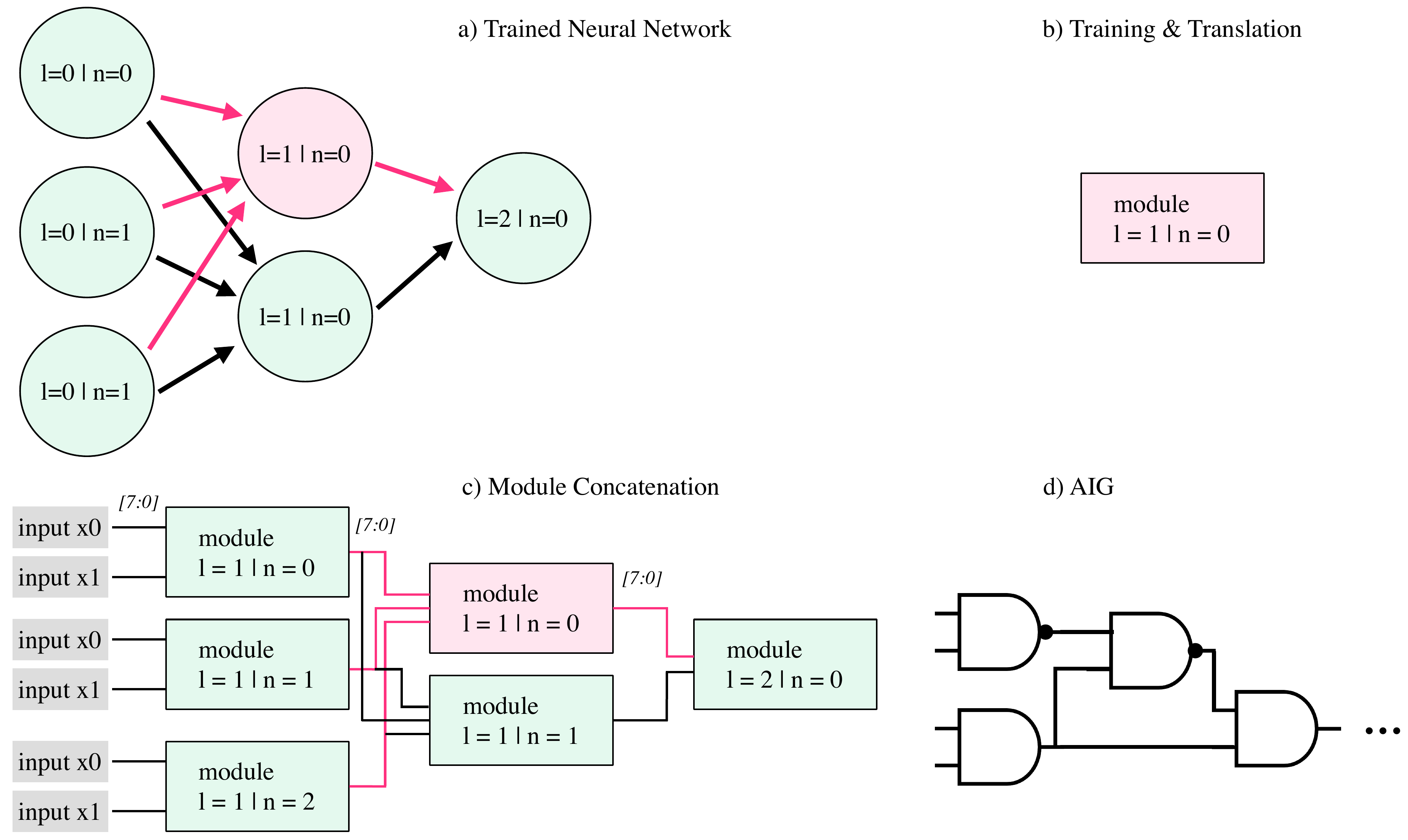}}
\caption{\small Illustration of translation from neural networks to logic \textbf{(a)} trained neural network, \textbf{(b)} intermediate input-output module, \textbf{(c)} contatenation of modules to recreate entire network, \textbf{(d)} translation to AIG.}
\label{framework_figure}
\end{center}
\vskip -0.2in
\end{figure*}

The following sections provide a detailed descriptions of the logic creation processes within the novel pipeline that we introduce. Figure \ref{framework_figure} further illustrates the pipeline as explained in Section \ref{framework}.

% \vspace{-3mm}
% \vspace{-3mm}

% -------------------------------------------------------
\subsubsection{Translation to an Arithmetic Circuit}
\label{nn_translation}
\begin{figure}[ht]
\begin{center}
\centerline{\includegraphics[width=0.8\textwidth]{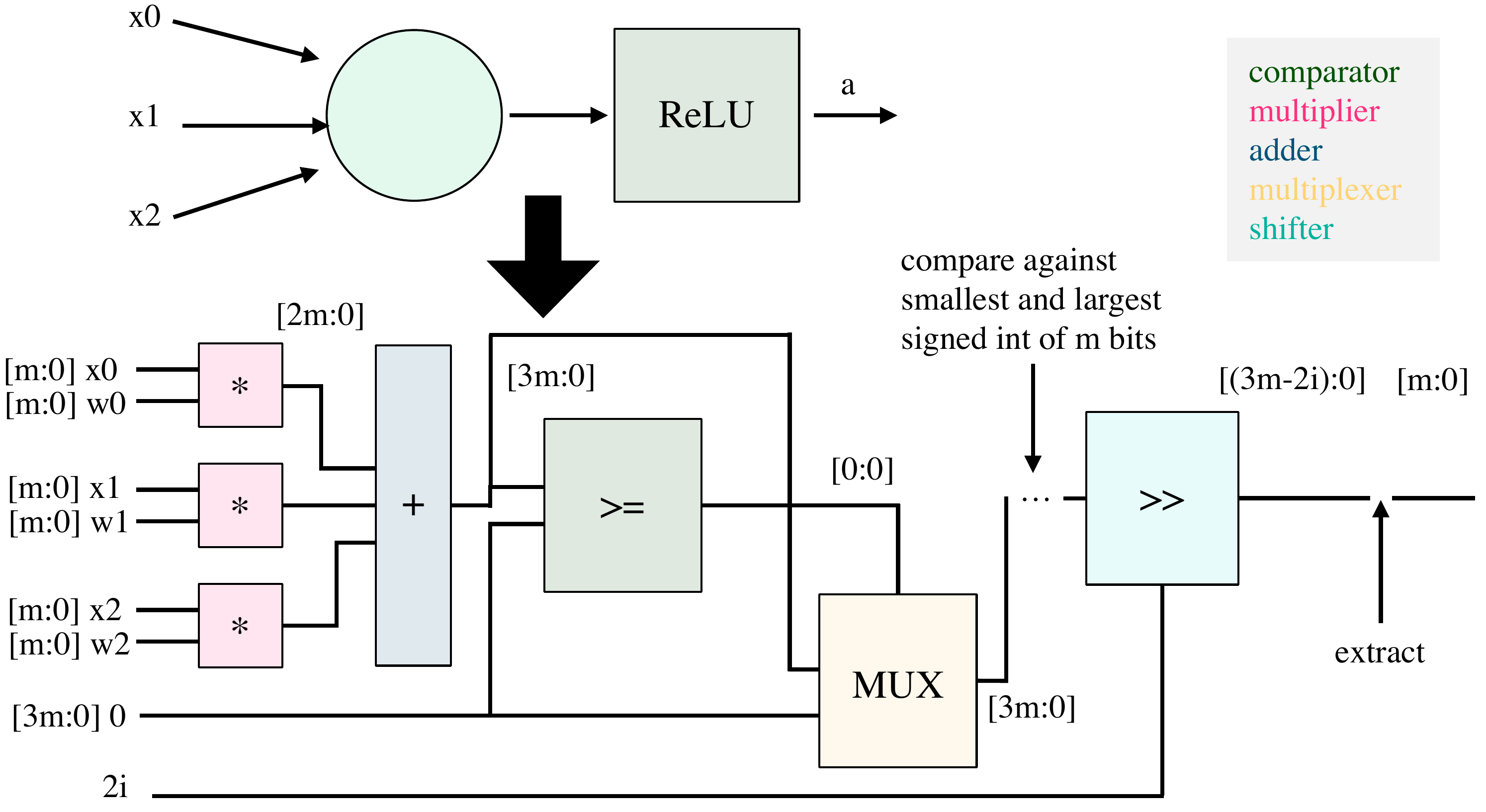}}
\caption{\small Overview of the \textbf{logic of a single neuron with a ReLU activation function} for a quantization scheme of $m$ bits. Multipliers are used for the weight-input multiplications and an accumulator is used for the addition. A comparator is used to compare against a constant input of $0$ to model the ReLU function and a shifter is used for a right shift to eliminate the additional $i$ fractional bits before extracting the $m$ least significant bits.}
\label{nn_translation_figure}
\end{center}
\end{figure}

Figure \ref{nn_translation_figure} shows how the operations during a forward pass at a neural network's node can be translated into logic. The multiplications ${w_i} \cdot {x_i}$ are executed by \emph{multipliers}. The two terms of the multiplication must have the same number of bits $m=k+i$ (with $k$ integer bits and $i$ fractional bits as used in the quantization scheme - Section \ref{quantization}). Due to the nature of binary multiplications, the result is of size $2m$ bits. The summation ${\sum}_i ({w_i} \cdot {x_i})$ is done by the \emph{accumulator logic}. To prevent an overflow, it is assumed to be of size $3m$ bits.

Then, a ReLU-activation function is modeled as \emph{comparator logic} with the $input_1$ of the accumulator and a constant $input_2$ of zero, both of size $3m$ bits. The comparator drives a \emph{multiplexer logic} that outputs $input_1$ if the comparator outputs $1$ (given by $input_1 > input_2$) and $input_2$ (meaning zero) otherwise. Finally, the result has to be brought back to size $m$.

This is done by a \emph{logical right shift} of $2i$ bits (to remove the additional fractional bits), followed by a clipping to the largest signed integer number and smallest signed integer number that can be represented with $m$ bits and hence, follows the procedure of the quantization scheme. After that, the result is still of size $3m - 2i$, but the relevant information stands at the $m$ least significant bits.

% -------------------------------------------------------
\subsubsection{Translation to Logic: Random Forest}
\label{rf_translation}

When training random forests on the activations of the neural network, the hyper-parameters that are used to influence the performance are the number of estimators (i.e. the number of decision trees in the forest) and the maximum depth (i.e. the maximum branch depth in a tree). Note that the depth of a decision tree influences the number of extracted features. Then each forest is turned separately into logic or to be more precise: each decision tree is turned into logic to provide its classification prediction. The result of each "logic-tree" is combined to form the prediction of the overall "logic-forest". Hence, it follows the same principle and operations as in "real-valued", standard decision trees. Nonetheless, there are small differences in how the random forests are trained and translated, which can have a high influence on the performance. Those differences are introduced in the following and are integrated into the pipeline in such a way that they can be seen as additional hyper-parameters, next to the number of estimators and the maximum tree depth. 

% -------------------------------------------------------
\paragraph{Inverse Weighting}\mbox{}\\
The random forests are sensitive to imbalanced training data sets. When training them on the activations of the neural network, it can happen that some activations that serve as labels are dominant over others. This occurs especially frequently for small quantization schemes, when small differences in the float values of activations are lost, due to falling into a small number of bins. When training random forests on such imbalanced data sets without further modifications, it can happen that during training they only select observations with the same label. This then leads to a constant class prediction of the most dominant class. That is problematic because the non-dominant class predictions are usually the ones that are of particularly high interest (e.g. in a binary classification problem with class 1 characterizing abnormality and class 0 for normality). To prevent this from happening, during the training process the samples of each class are weighted inversely according to the class label distribution. That means that observations of less dominant classes have a higher chance to be chosen and influence the feature extraction process. It is referred to this procedure in the following as \emph{inverse weighting}. To summarize, it is something that influences the training process of random forests to overcome class imbalances, but does not influence the logic translation itself.

% -------------------------------------------------------
\paragraph{Logic Translation of a Single Decision Tree - Main Principle}\mbox{}\\
The final leaves of the decision tree implementations model class probabilities for each of the class labels, according to the distribution of observed training examples that fell into the node's bin. Instead of traversing the tree from top to bottom and either going left or right at each branch according to the decision rule, in the logic the tree is propagated backwards from bottom to top. This is done for the reason that a final wire is needed at which the signal, i.e. final class prediction, can be propagated to the next logic module. The simplest way to do so is the root node. Hence, the decision rules are evaluated the opposite way and the class probabilities are propagated upwards to the root node according to the decision rules at the branches. This means that a cascade of comparators does the job of evaluating the decision rules and drives a cascade of multiplexers which select the corresponding class probabilities that need to be propagated. Thus, when executing the logic as simulation on quantized test signals, the root node's wires receive signals that model the class probabilities of the tree's node at which one lands when following the learned decision rules. This is the main principle of translating a single decision tree to logic. However, there are detailed differences in what is propagated and how the predictions of multiple trees are handled in the logic to form a whole prediction of the forest. The differences are dependent on whether a bitwise or none-bitwise option of logic creation is chosen. This is explained in the following. 

\begin{figure}[ht]
\begin{center}
\centerline{\includegraphics[width=0.8\textwidth]{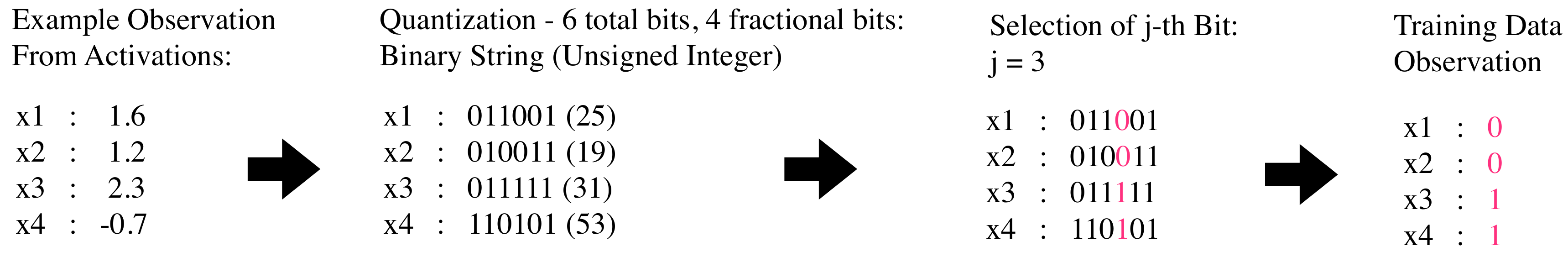}}
\caption{\small Example of how the \textbf{$j^{th}$ bit of each quantized activation is selected} to form the features for \textbf{bitwise training} and logic translation.}
\label{bitwise_training_bitselection}
\end{center}
\end{figure}

% -------------------------------------------------------
\paragraph{Bitwise Training}\mbox{}\\
An option that influences both the training and the translation is the option for \emph{bitwise training}. Within that, the neural networks activations are quantized to binary strings. Then, a random forest is trained on each bit $0{\leq}j<m$ separately and solves a binary classification problem. Note that subsets of the quantized activation data have to be formed that always extract the $j^{th}$ bit of each feature and label. Thus, the decision trees also receive samples with each feature being a single-bit, as illustrated in Figure \ref{bitwise_training_bitselection}. This procedure is also done for simplifying the logic itself. When only using binary data, the random forest problem is simplified to thresholds of $0.5$ at the branches as there are only two possible values $0$ and $1$ (or any other fixed threshold between 0 and 1). What remains to be learned by the random forests is which feature is relevant for the classification. This is equivalent to learning a removal of ``unimportant" variables and hence, to a don't-care-based dependency elimination that might be able to drive generalization as described in Section \ref{dont-care-elimination}.

\begin{figure}[ht]
\begin{center}
\centerline{\includegraphics[width=0.75\textwidth]{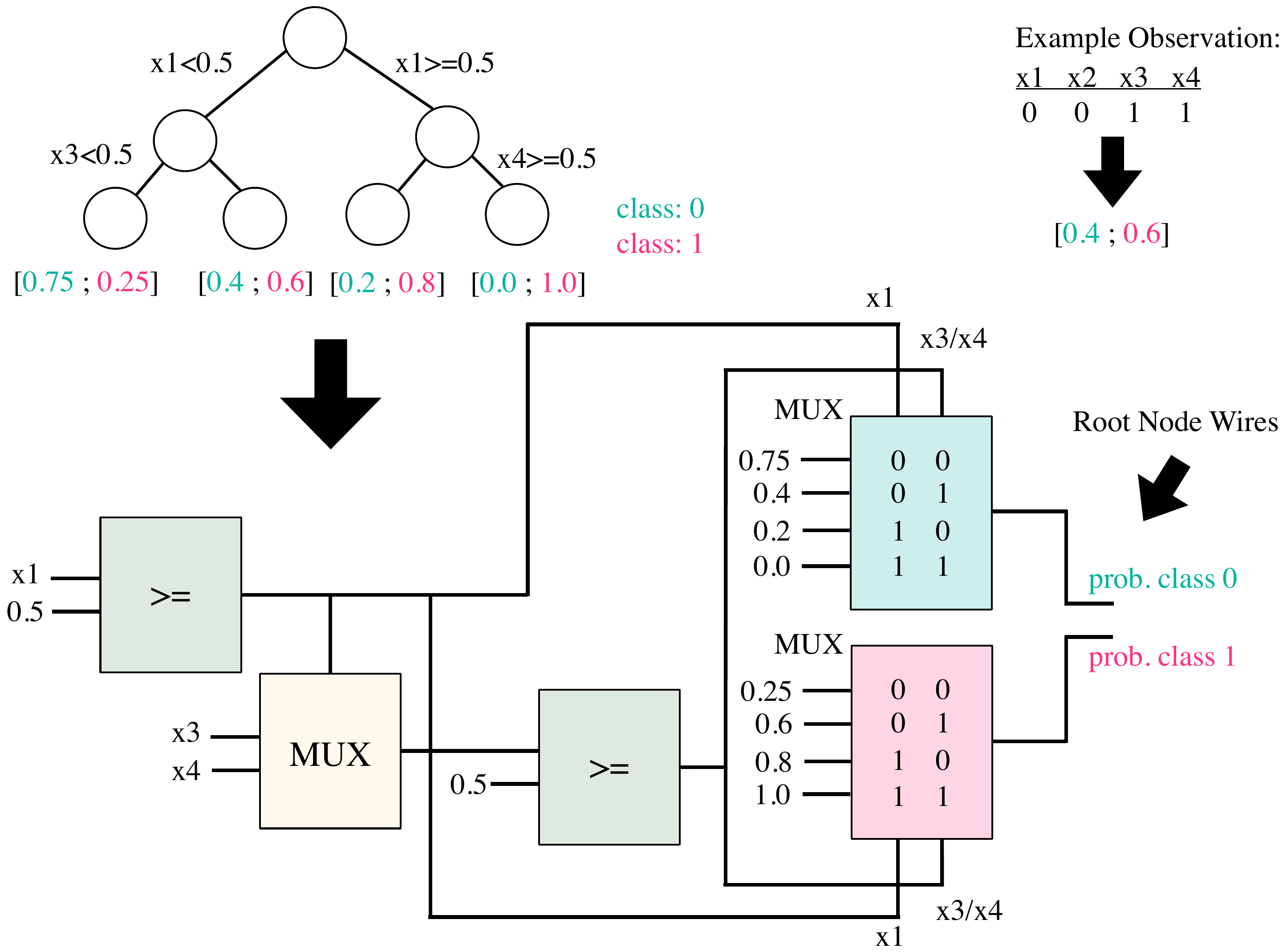}}
\caption[Logic Translation of a Single Decision Tree - Bitwise Training]{Overview of the \textbf{logic of a single decision tree} of depth 2 for 4 data features under \textbf{bitwise training}. The final leaves model class probabilities of the binary classification problem. The logic is implemented as a cascade of comparators and multiplexers.}
\label{bitwise_decisiontree_figure}
\end{center}
\end{figure}

In this setting, the probabilities of a prediction whether the bit should be classified as 1 or 0 are propagated. Hence, the root node of a decision tree for a single-bit classification has two wires of $m$ bits that model the probability of the sample being class 0 and class 1. The logic of an exemplary single decision tree under bitwise training is visualized in Figure \ref{bitwise_training_bitselection}. \\

The probability wires at each decision tree's root node are used to form overall class probabilities of the random forest. This is done by summing up all propagated probabilities of class 0 and class 1 across the trees with an accumulator circuit. Same as in the setting of the arithmetic circuit (Section \ref{nn_translation}) the accumulator is assumed to be of size $3m$ bits to prevent overflow. In the following it is referred to this summing up operation as a \emph{majority vote} because each decision tree contributes equally to the class probabilities. \\

\begin{figure}[ht]
\begin{center}
\centerline{\includegraphics[width=0.5\textwidth]{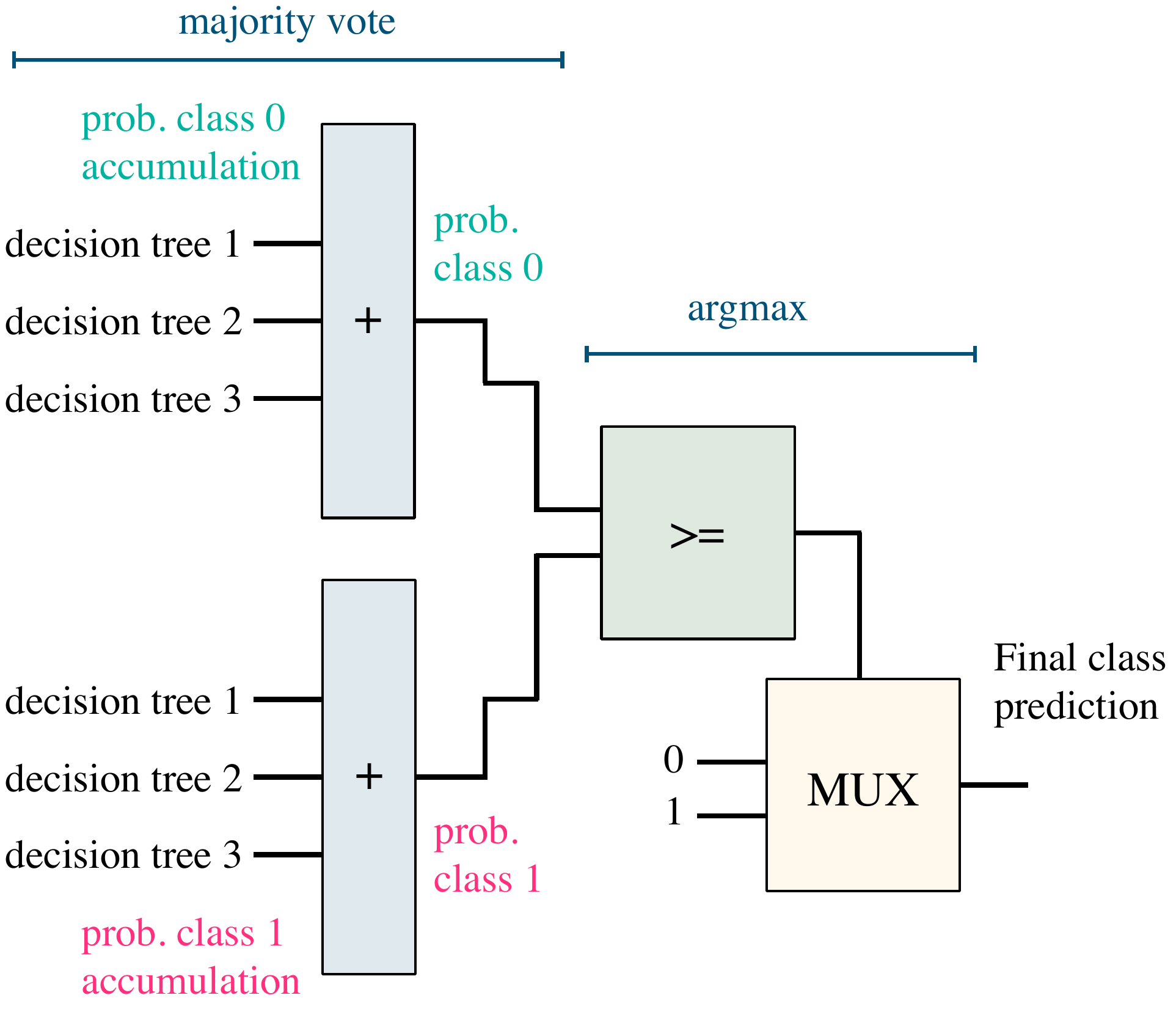}}
\caption{\small Overview of the \textbf{majority vote and argmax logic of a random forest} under \textbf{bitwise training}. The example contains 3 decision trees, where the class probabilities that were propagated to the root nodes (see Figure \ref{bitwise_decisiontree_figure}) serve as input.}
\label{bitwise_randomforest_figure}
\end{center}
\end{figure}

To then propagate the actual single-bit class prediction of the forest, a logic operation follows that mimics the argmax-function via a comparator that compares the summed-up probabilities. The comparator drives a multiplexer which chooses between a constant 1-bit signal of 1 or 0. To form the final signal that is provided as input to the next logic module, all single bit predictions of the $m$ forests are again concatenated to form the $m$-bit signal. Both the majority vote and the argmax operation are visualized in Figure \ref{bitwise_randomforest_figure}.

% -------------------------------------------------------
\paragraph{Influence of the Argmax-Function in the Bitwise Training Setting}\mbox{}\\
It should be highlighted that the definition of the argmax-function can influence the class prediction in such single-bit random forest logic settings. In a usual real-valued setting it is not possible that the probability $p(\text{class 1})$ equals the probability $p(\text{class 0})$ because they are dependent on each other through:  

\begin{equation}
p(\text{class 1}) = 1 - p(\text{class 0})
\end{equation}

However, as the probabilities are summed up across multiple decision trees to form "pseudo-probabilities", it can happen that they are equal. This would mean that also the quantized binary strings that go into the comparator would be equal. Especially, it can also happen that those binarized signals are equal, although the real-valued, summed-up probabilities are unequal. The reason for this can be a small quantization scheme, where e.g.  $p_{\text{sum}}(\text{class 0})=0.631$ and $p_{\text{sum}}(\text{class 1})=0.632$ lead to the same quantized value and cannot be further distinguished. When such identical signals occur it makes a difference whether the argmax is defined as in Equation \ref{argmax_def1}, which favours and chooses class 1, or as in Equation \ref{argmax_def0}, where class 0 is favoured and chosen. Nonetheless, it is supposed that the cases where the implementation matters are rather neglectable, although they in theory occur more frequently when choosing small quantization schemes. The influence is not further investigated in the following and Equation \ref{argmax_def1} is used as definition of the argmax in this bit-wise random forest logic setting throughout the rest of the thesis. 

\begin{equation}
\label{argmax_def1}
\underset{\{ \text{class 0; class 1} \}}{argmax} = 
 \begin{cases}
    1 & \text{if } p(\text{class 1}) \geq p(\text{class 0})\\
    0 & \text{otherwise}
\end{cases}
\end{equation}

\begin{equation}
\label{argmax_def0}
\underset{\{ \text{class 0; class 1} \}}{argmax} = 
 \begin{cases}
    1 & \text{if } p(\text{class 1}) > p(\text{class 0})\\
    0 & \text{otherwise}
\end{cases}
\end{equation}

% -------------------------------------------------------
\begin{figure}[ht!]
\begin{center}
\centerline{\includegraphics[width=0.7\textwidth]{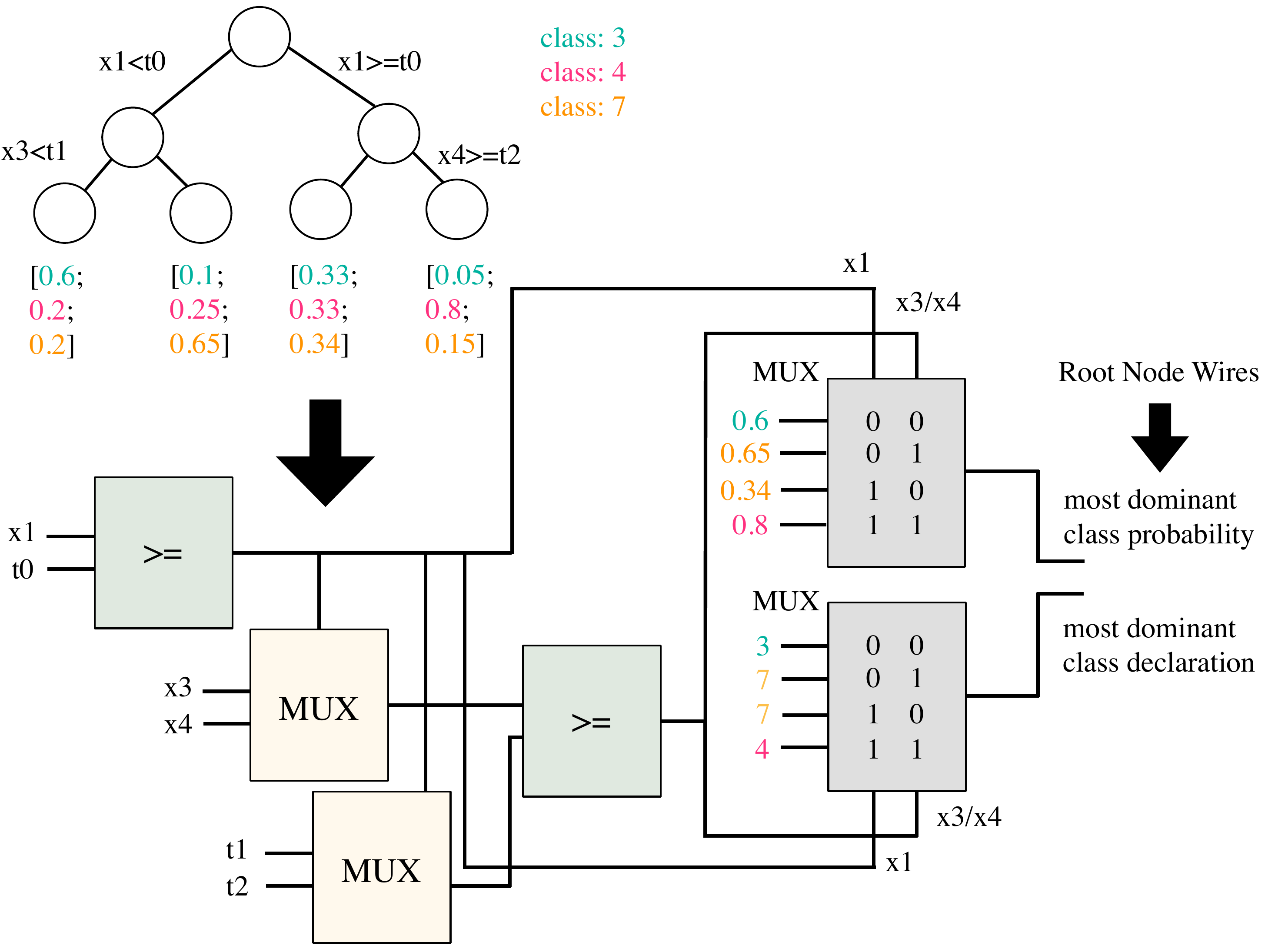}}
\caption{\small Overview of the \textbf{logic of a single decision tree} of depth 2 for 4 data features under \textbf{none-bitwise training}. In this example the quantization scheme has $m=3$ total bits, hence 8 possible classes, but only three classes are observed in the training data set. Only the most dominant class and its probability is propagated from each leaf node to the root node. Note that also the thresholds at the branches have to be learned and have to be selected in the logic via multiplexers.}
\label{nonebitwise_decisiontree_figure}
\end{center}
\end{figure}

\paragraph{None-Bitwise Training}\mbox{}\\
In contrast to the bitwise training, the \emph{none-bitwise option} does not train $m$ random forests for a quantization scheme of $m$ total bits and does not operate on only binary features and labels. Instead of using the binary strings of an observation's features, the unsigned integer representation of each feature is used from the quantization scheme. Hence, just one random forest is trained on integer values of the quantized activations and besides learning relevant features for the decision trees, also the thresholds need to be learned. However, for a quantization scheme of $m$ total bits, this means that $2^m$ classes could occur to be learned by the decision tree. Propagating all class probabilities at each of a tree's branches is practically infeasible and would lead to a complex wiring and to a high number of comparisons. Therefore, at each branch a comparator only compares the (propagated) probabilities of the left and right nodes' most dominant class predictions. The comparator again drives a multiplexer that selects the class which corresponds to the higher probability. Hence, the decision tree's root node has only two wires at the end: one for the propagated class prediction and one for its probability. No logic implementation of the argmax is needed. Figure \ref{nonebitwise_decisiontree_figure} shows an example of the logic of a single decision tree under none-bitwise training with $m=3$ total bits, hence $2^3 = 8$ possible classes, but only three classes being observed in the training data set. \\

\begin{figure}[ht]
\begin{center}
\centerline{\includegraphics[width=0.7\textwidth]{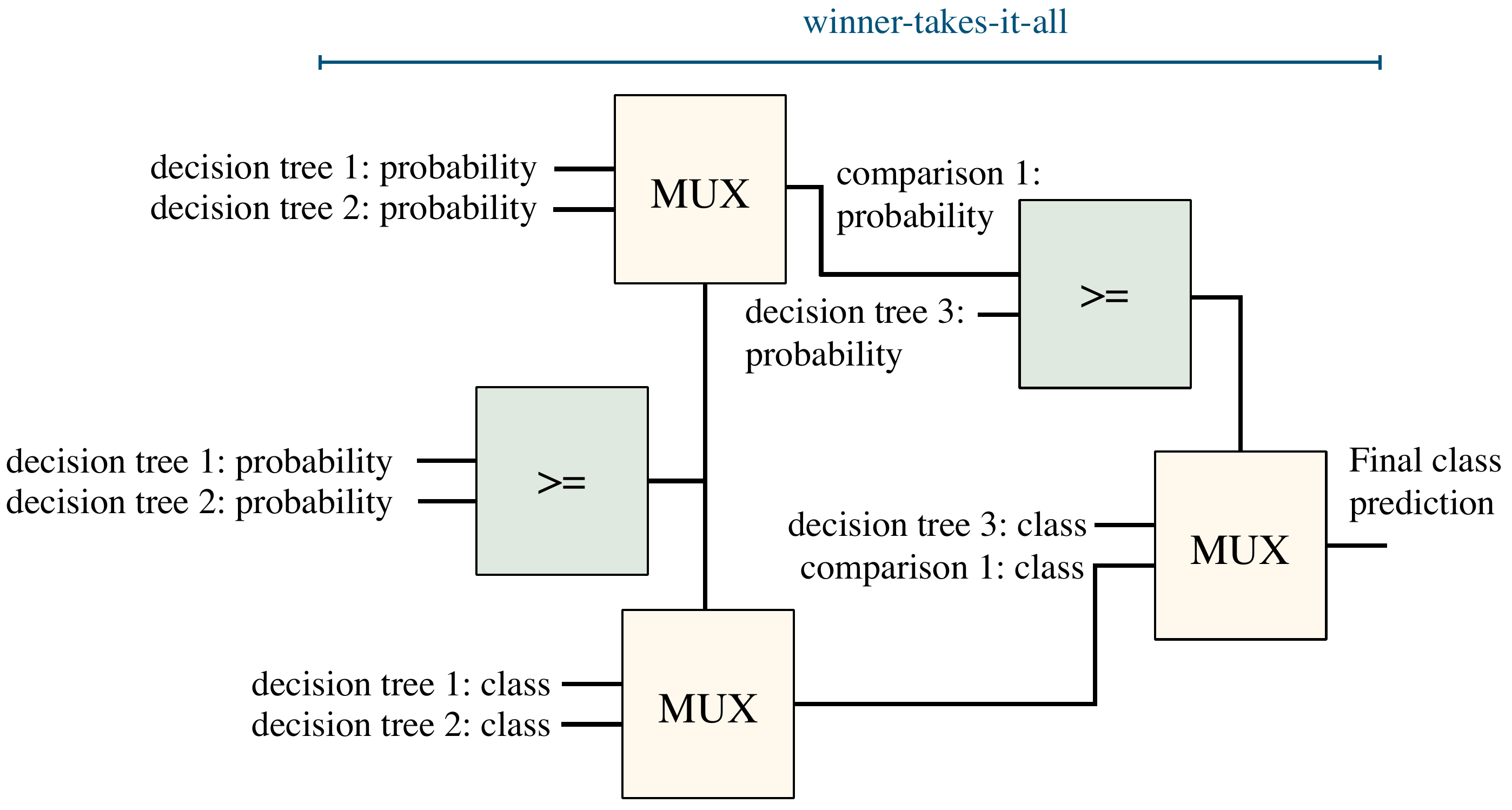}}
\caption{\small Example of the \textbf{winner-takes-it-all logic of a random forest} under \textbf{none-bitwise training}. The example contains 3 decision trees, where the class declarations and corresponding probabilities that were propagated to the root nodes (see Figure \ref{nonebitwise_decisiontree_figure}) serve as input.}
\label{nonebitwise_randomforest_figure}
\end{center}
\end{figure}

To receive the final class prediction of the overall random forest in logic, also the majority vote needs to be replaced. The decision trees' root nodes have varying class predictions and thus, summing up class probabilities across trees is not possible anymore. Instead, a comparator is again used to drive a multiplexer that selects the class prediction corresponding to the highest probability across all root nodes. In the following, it is referred to this procedure as \emph{winner-takes-it-all} concept because a single decision tree that is the most certain about its prediction delivers the classification result of the whole random forest. The logic implementation of the winner-takes-it-all concept is visualized in Figure \ref{nonebitwise_randomforest_figure}.

% -------------------------------------------------------
\subsubsection{Translation to Logic: LogicNet}                               
\label{lgn_translation}
\begin{figure}[ht]
\begin{center}
\centerline{\includegraphics[width=0.6\textwidth]{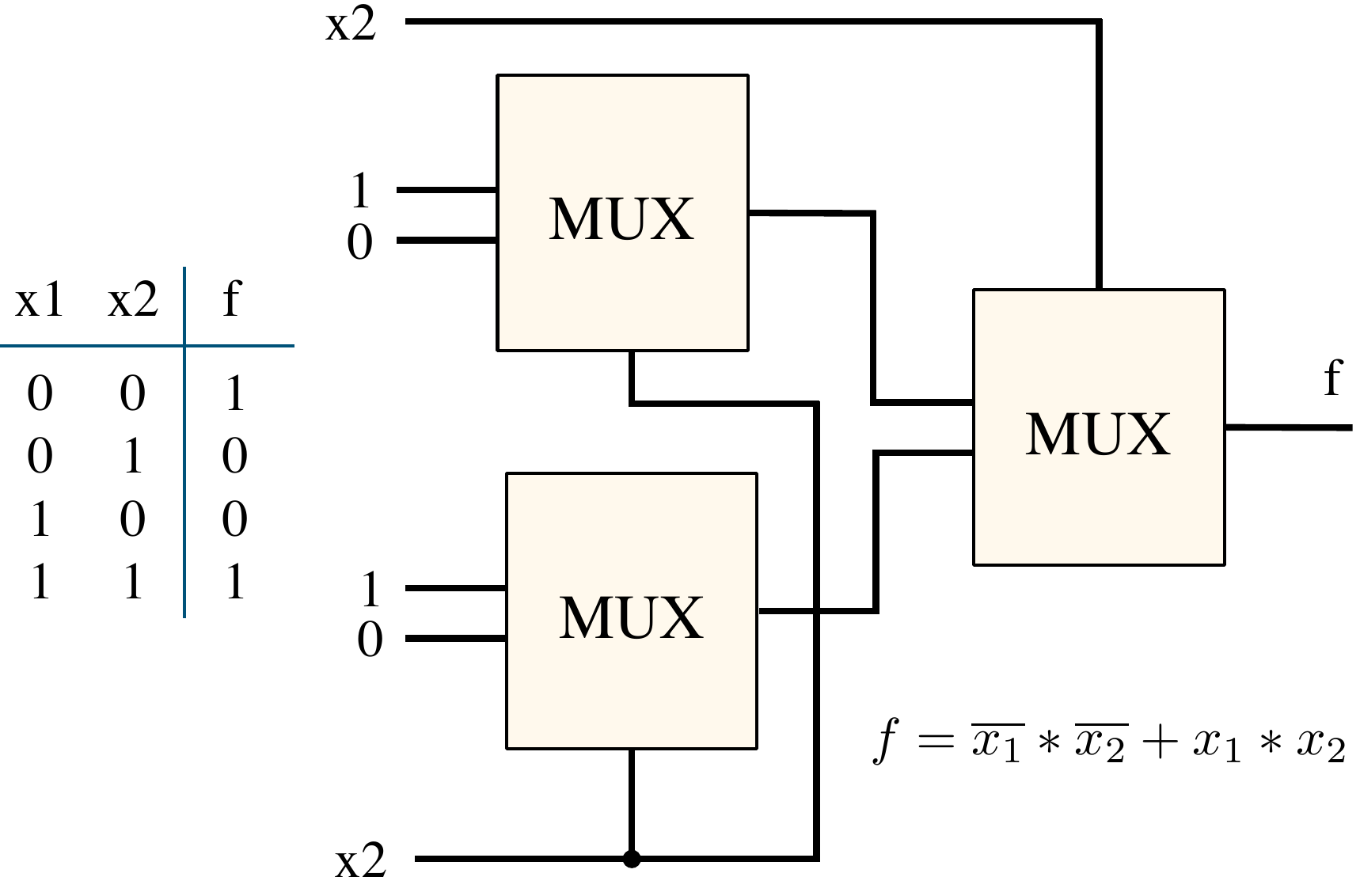}}
\caption[Logic Translation of a LUT]{Example of a \textbf{look-up table (LUT) logic implementation} with multiplexers for a given Boolean function. Such LUTs are stacked in multiple layers to form LogicNet.}
\label{lut_figure}
\end{center}
\end{figure}

Similar to the bitwise training of random forests, LogicNets are trained on binary classifications (i.e. one LogicNet per bit $0{\leq}j{<}m$) which are translated into logic separately, as also visualized in Figure \ref{bitwise_training_bitselection}. As LogicNet only consists of look-up-tables, the simplest way of implementation is with a cascade of multiplexers, as visualized in Figure \ref{lut_figure}. The logic of each LogicNet to model node $l;n$ is concatenated to one module to form $LGN_{l;n}^{logic}$ (Overview - Step 5 in Section \ref{framework}) with $N_{l-1}$ inputs of $m$ bits and an output of $m$ bits. The implementation of LogicNet is taken from \cite{learningandmem} and only supports binary classifications. Introducing a none-bitwise training option as for the random forests is left as future work.

% -------------------------------------------------------
\subsubsection{Last Layer Logic}
\label{last_layer_logic}
\begin{figure}[ht]
\begin{center}
\centerline{\includegraphics[width=0.7\textwidth]{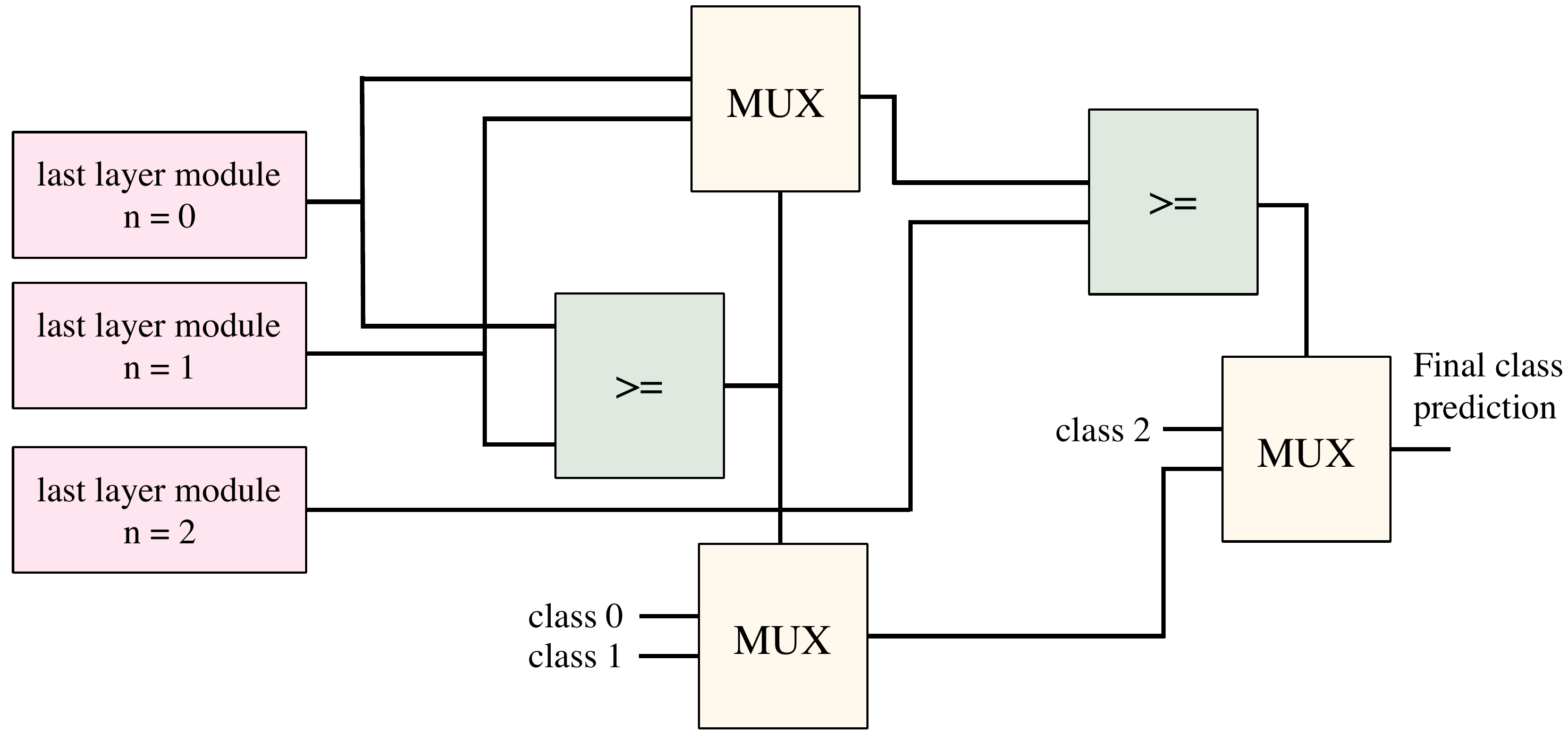}}
\caption{\small Example of the \textbf{last layer logic} argmax-implementation for a 3-class classification problem, where the last layer modules serve as input.}
\label{last_layer_logic_figure}
\end{center}
\end{figure}

Independent of the logic translation method (arithmetic circuit, random forest and LogicNet), an additional logic block has to be appended to the last layer's logic module, in order to form the final class prediction. Recall that the logic modules that simulate the last layer of the neural network model the activations that come from a softmax function. There are as many nodes in the neural network's last layer (and thus, also as many logic modules) as there are classes in the classification task. To derive the class prediction from softmax values in a real-valued setting, the class would be chosen that corresponds to the node with the highest softmax value. This is what needs to be replaced in the logic by a comparator and multiplexer logic that does the job of an argmax, as similarly described in Section \ref{rf_translation} and as illustrated in Figure \ref{last_layer_logic_figure}. Note the similarity to the winner-takes-it-all concept. The class class is chosen that corresponds to the last layer's logic module with the greatest binary signal.

% --------------------------------------------------------------
\subsection{Full Results}
This section provides extensive tables for all training setups and configurations of the pipeline. The tables report the mean and standard deviations for each setup being executed with three independent runs and the result of the top-1 model of all runs. Additional horizontal lines are drawn in the tables to clearly mark the separation between the top-1 models performing better than, equal to or worse than random.

% -------------------------------------------------------------
\clearpage
\subsection{Arithmetic Circuit Baselines}

The results for translating the real-valued neural network baselines from Table \ref{real_nn_baselines_refined} into arithmetic circuits under different quantization schemes are given in the following. 
\vspace{-3mm}
\begin{table}[ht!]
\caption{\small The results of the \textbf{"GIB Model 1" arithmetic circuit}.}
\label{gib_arithmetic_circuit}
\centering
\begin{adjustbox}{width=0.79\textwidth}
\begin{tabular}{ccccccc}
\toprule
\makecell{Total\\Bits} & \makecell{Fractional\\Bits} & \makecell{AIG\\Nodes} &    \makecell{AIG\\Levels} &  \makecell{Accuracy\\in \%} & \makecell{Real-Valued Accuracy\\(Preserved) in \%} \\
\midrule
18		&              10	&          123,389 &           451 &  \textbf{83.28} 	& 83.28 (100.0)	\\
32		&              16	&          357,662 &           672 &           83.28 	& 83.28 (100.0)	\\
16		&              10	&          107,068 &           419 &           83.11 	& 83.28 (99.79)	\\
 7		&               3	&  \textbf{14,314} &         \textbf{231} &           70.36 	& 83.28 (84.49)	\\
10		&               6	&           42,051 &           312 &            64.40 	& 83.28 (77.34)	\\
12		&               8	&           64,535 &           359 &           63.25 	& 83.28 (75.95)	\\
6		&               4	&           16,436 &           235 &            50.50	& 83.28 (60.63)	\\
\midrule
8		&               6	&           32,538 &           286 &            50.00 	& 83.28 (60.04)	\\
\bottomrule
\end{tabular}
\end{adjustbox}
\end{table}
\vspace{-3mm}

\begin{table}[ht]
\caption{\small The results of the \textbf{"GIB Model 2" arithmetic circuit}.}
\label{gib_nn_logic}
\centering
\begin{adjustbox}{width=0.79\textwidth}
\begin{tabular}{cccccc}
\toprule
\makecell{Total\\Bits} & \makecell{Fractional\\Bits} & \makecell{AIG\\Nodes} & \makecell{AIG\\Levels} &  \makecell{Accuracy\\in \%} & \makecell{Real-Valued Accuracy\\(Preserved) in \%} \\
\midrule
         6 &               4 &           38,414 &           240 &  \textbf{52.48} 	& 85.93 (61.07)\\
         8 &               6 &           77,059 &           290 &           51.49 	& 85.93 (59.89)\\
\midrule
        12 &               8 &          150,427 &           338 &            45.20 	& 85.93 (52.60) \\
        10 &               6 &           95,582 &           312 &           43.05 	& 85.93 (50.09) \\
         7 &               3 &  \textbf{33,557} &  \textbf{238} &         35.93 	& 85.93 (41.81) \\
        16 &              10 &          248,322 &           392 &            29.30 	& 85.93 (34.09) \\
        18 &              10 &          281,471 &           416 &            29.30 	& 85.93 (34.09) \\
        32 &              16 &          823,495 &           592 &            29.30 	& 85.93 (34.09) \\
\bottomrule
\end{tabular}
\end{adjustbox}
\end{table}
\vspace{-3mm}

\begin{table}[ht!]
\caption{\small The results of the \textbf{"VACS Model 1" arithmetic circuit}.}
\label{vacs_arithmetic_circuit}
\centering
\begin{adjustbox}{width=0.79\textwidth}
\begin{tabular}{ccccccc}
\toprule
\makecell{Total\\Bits} & \makecell{Fractional\\Bits} & \makecell{AIG\\Nodes} &    \makecell{AIG\\Levels} &  \makecell{Accuracy\\in \%} & \makecell{Real-Valued Accuracy\\(Preserved) in \%} \\
\midrule
32 	&              16 	&          338,757 		&           549 	&  \textbf{63.20} 	& 63.20 (100.0) \\
18 	&              10 	&          119,946 		&           348 	&          62.59 		& 63.20 (99.03)\\
16 	&              10 	&          104,357 		&           320 	&          51.56 		& 63.20 (81.58) \\
12 	&               8 	&           63,560 		&           266 	&          36.12 		& 63.20 (57.15) \\
7 	&               3 	&           17,807 		&  \textbf{183} 	&          35.51 		& 63.20 (56.19) \\
10 	&               6 	&           43,612 		&           233 	&          34.35 		& 63.20 (54.35) \\
\midrule
6 	&               4 	&  \textbf{17,507} 		&           184 	&          33.88 		& 63.20 (53.61) \\
8 	&               6 	&           31,102 		&           221 	&          33.13 		& 63.20 (52.42) \\
\bottomrule
\end{tabular}
\end{adjustbox}
\end{table}
\vspace{-3mm}

\begin{table}[ht!]
\caption{\small The results of the \textbf{"VACS Model 2" arithmetic circuit}.}
\label{vacs_nn_logic_big}
\centering
\begin{adjustbox}{width=0.79\textwidth}
\begin{tabular}{cccccc}
\toprule
\makecell{Total\\Bits} & \makecell{Fractional\\Bits} & \makecell{AIG\\Nodes} & \makecell{AIG\\Levels} &  \makecell{Accuracy\\in \%} & \makecell{Real-Valued Accuracy\\(Preserved) in \%} \\
\midrule
         6 &               4 &  \textbf{106,572} &  \textbf{353} &  \textbf{34.97} 	& 75.58 (46.26) \\
        32 &              16 &          2,173,117 &           726 &           34.97 		& 75.58 (46.26)\\
         7 &               3 &           109,473 &           362 &           33.54  		& 75.58 (44.37)\\
        10 &               6 &           270,088 &           414 &           33.54 		& 75.58 (44.37) \\
        18 &              10 &           755,160 &           522 &            33.40		& 75.58 (44.19)  \\
\midrule
        16 &              10 &           668,259 &           496 &           32.93		& 75.58 (43.57)  \\
         8 &               6 &           195,020 &           387 &           32.45		& 75.58 (42.93)  \\
        12 &               8 &           411,030 &           444 &           30.27		& 75.58 (40.05)  \\
\bottomrule
\end{tabular}
\end{adjustbox}
\end{table}

% -------------------------------------------------------------
% \newpage
\clearpage
\subsubsection{Random Forest Evaluation}
The tables provided in the following show the results of the logic creation with random forests as intermediate steps. They are based on the real-valued neural network baselines defined in Table \ref{real_nn_baselines_refined} that served as input to the logic pipeline. In each table, the results are sorted in descending manner according to the top-1 model's performance and the column-wise "winner" is marked in bold.

Note that if the number of nodes or levels in the AIG is zero, this means that the random forests have learned nothing and its corresponding logic has a constant class prediction. 

% --------------------------------------------------------------
% \newpage
%\clearpage
\begin{table}[ht!]
\centering
\caption{\small Results of the \textbf{random forest logic} on the \textbf{"GIB - Model 1"} setup for a quantization scheme of \textbf{7 total bits and 3 fractional bits}.}\label{rf_tb7_gib_results}
\begin{adjustbox}{width=\textwidth}
\begin{tabular}{ccccccccc}
\toprule
\makecell{Estimators} & \makecell{Max.\\Depth} & \makecell{Bitwise} & \makecell{AIG Nodes\\(Top1)} & \makecell{AIG Nodes\\(TopN)} & \makecell{AIG Levels\\(Top1)} & \makecell{AIG Levels\\(TopN)} & \makecell{Accuracy\\(Top-1)} & \makecell{Accuracy\\(Top-N)} \\
\midrule
3 &	20 &      No &           50,684 &   50,817.67 (199.52) &               245 &     245.00 (0.00) &  \textbf{83.28} &    \textbf{83.28 (0.00)} \\
2 &	20 &      No &           37,785 &   37,909.67 (206.48) &               219 &     219.00 (0.00) &           83.11 &    82.95 (0.28) \\
4 &	10 &      No &           32,344 &     32,344.00 (0.00) &               255 &     255.00 (0.00) &           82.28 &    82.28 (0.00) \\
2 &	10 &      No &           21,340 &     21,340.00 (0.00) &               201 &     201.00 (0.00) &           82.28 &    82.28 (0.00) \\
2 & 	15 &      No &           35,270 &    35,159.33 (99.05) &               215 &     215.00 (0.00) &           82.12 &    81.73 (0.34) \\
4 &	5 &      No &           13,654 &     13,654.00 (0.00) &               233 &     233.00 (0.00) &           82.12 &    82.12 (0.00) \\
3 &	10 &      No &           26,578 &     26,578.00 (0.00) &               228 &     228.00 (0.00) &           81.95 &    81.95 (0.00) \\
\bottomrule
\end{tabular}
\end{adjustbox}
\end{table}

% --------------------------------------------------------------
%\newpage
\begin{table}[ht!]
\centering
\caption{\small Results of the \textbf{random forest logic} on the \textbf{"GIB Model 1"} setup for a quantization scheme of \textbf{18 total bits and 10 fractional bits}.}\label{rf_tb18_gib_results}
\begin{adjustbox}{width=\textwidth}
\begin{tabular}{ccccccccc}
\toprule
\makecell{Estimators} & \makecell{Max.\\Depth} & \makecell{Bitwise} & \makecell{AIG Nodes\\(Top1)} & \makecell{AIG Nodes\\(TopN)} & \makecell{AIG Levels\\(Top-1)} & \makecell{AIG Levels\\(TopN)} & \makecell{Accuracy\\(Top1)} & \makecell{Accuracy\\(Top-N)} \\
\midrule
2 &         10 &      No &          114,011 &      114,011.00 (0.00) &               357 &     357.00 (0.00) &  \textbf{77.65} &    \textbf{77.65 (0.00)} \\
3 &         10 &      No &          122,839 &      122,839.00 (0.00) &               426 &     426.00 (0.00) &           76.66 &    76.66 (0.00) \\
2 &         15 &      No &          142,875 &      142,875.00 (0.00) &               381 &     381.00 (0.00) &           60.76 &    60.76 (0.00) \\
4 &         15 &     Yes &          146,125 &   147,417.33 (2560.10) &               328 &     328.33 (0.58) &           58.28 &    51.38 (6.44) \\
4 &         20 &      No &          213,640 &      213,639.33 (1.15) &               542 &     542.00 (0.00) &           57.78 &    57.78 (0.00) \\
4 &         10 &      No &          129,579 &      129,579.67 (0.58) &               495 &     495.00 (0.00) &           57.45 &    57.45 (0.00) \\
3 &         20 &      No &          189,521 &      189,521.00 (0.00) &               474 &     474.00 (0.00) &           57.28 &    57.28 (0.00) \\
\bottomrule
\end{tabular}
\end{adjustbox}
\end{table}

% --------------------------------------------------------------
%\newpage
\begin{table}[ht!]
\centering
\caption{\small Results of the \textbf{random forest logic} on the \textbf{"GIB Model 2"} setup for a quantization scheme of \textbf{6 total bits and 4 fractional bits}.}\label{rf_tb6_gib_big_results}
\begin{adjustbox}{width=\textwidth}
\begin{tabular}{ccccccccc}
\toprule
\makecell{Estimators} & \makecell{Max.\\Depth} & \makecell{Bitwise} & \makecell{AIG Nodes\\(Top1)} & \makecell{AIG Nodes\\(TopN)} & \makecell{AIG Levels\\(Top1)} & \makecell{AIG Levels\\(TopN)} & \makecell{Accuracy\\(Top-1)} & \makecell{Accuracy\\(Top-N)} \\
\midrule
3 &         15 &     Yes &          425,617 &  429,633.67 (3740.06) &               213 &     213.00 (0.00) &  \textbf{56.29} &    50.33 (5.42) \\
2 &          5 &     Yes &           37,935 &    381,17.67 (333.86) &      \textbf{174} &     \textbf{174.00 (0.00)} &           55.46 &    52.59 (2.50) \\
4 &          5 &     Yes &           54,120 &    54,278.00 (136.84) &               198 &     198.00 (0.00) &            55.30 &    \textbf{52.98 (2.40)} \\
2 &         20 &     Yes &          295,923 &   295,516.33 (756.92) &               201 &     201.00 (0.00) &            54.80 &    49.61 (4.73) \\
3 &          5 &     Yes &           46,260 &    46,134.67 (116.38) &               186 &     186.00 (0.00) &           53.81 &    51.99 (2.01) \\
4 &         15 &     Yes &          567,076 &  559,755.00 (6486.18) &               225 &     225.00 (0.00) &           52.81 &    50.28 (3.44) \\
4 &         20 &     Yes &          557,274 &  558,920.67 (1602.50) &               225 &     225.00 (0.00) &           52.48 &    49.23 (2.85) \\
\bottomrule
\end{tabular}
\end{adjustbox}
\end{table}

% --------------------------------------------------------------
%\newpage
\begin{table}[ht!]
\centering
\caption{\small Results of the \textbf{random forest logic} on the \textbf{"GIB Model 2"} setup for a quantization scheme of \textbf{12 total bits and 8 fractional bits}.}\label{rf_tb12_gib_big_results}
\begin{adjustbox}{width=\textwidth}
\begin{tabular}{ccccccccc}
\toprule
\makecell{Estimators} & \makecell{Max.\\Depth} & \makecell{Bitwise} & \makecell{AIG Nodes\\(Top1)} & \makecell{AIG Nodes\\(TopN)} & \makecell{AIG Levels\\(Top1)} & \makecell{AIG Levels\\(TopN)} & \makecell{Accuracy\\(Top-1)} & \makecell{Accuracy\\(Top-N)} \\
\midrule
2 &         20 &      No &           214,009 &    214,352.00 (297.05) &               300 &     300.00 (0.00) &  \textbf{59.27} &    \textbf{59.27 (0.00)} \\
4 &         10 &     Yes &          1,135,087 &  1,132,649.00 (3201.05) &               287 &     287.00 (0.00) &           57.12 &    51.60 (5.91) \\
2 &          5 &     Yes &           155,585 &    155,780.67 (437.64) &      \textbf{242} &     \textbf{241.67 (0.58)} &           56.29 &    53.15 (2.82) \\
2 &         20 &     Yes &           704,065 &   710,442.67 (5990.50) &               267 &     267.00 (0.00) &            54.30 &    51.76 (2.30) \\
4 &         15 &     Yes &          1,301,420 &   1,300,616.67 (941.93) &               291 &     291.00 (0.00) &           53.31 &    52.37 (1.00) \\
3 &         10 &     Yes &           877,825 &   879,918.00 (2582.49) &               275 &     275.00 (0.00) &           52.98 &    51.05 (2.30) \\
3 &          5 &     Yes &           186,340 &    186,607.00 (355.86) &               253 &     253.33 (0.58) &           52.15 &    47.79 (3.82) \\
\bottomrule
\end{tabular}
\end{adjustbox}
\end{table}

% --------------------------------------------------------------
%\newpage
\begin{table}[ht!]
\centering
\caption{\small Results of the \textbf{random forest logic} on the \textbf{"VACS Model 1"} setup for a quantization scheme of \textbf{7 total bits and 3 fractional bits}.}\label{rf_tb7_vacs_results}
\begin{adjustbox}{width=\textwidth}
\begin{tabular}{ccccccccc}
\toprule
\makecell{Estimators} & \makecell{Max.\\Depth} & \makecell{Bitwise} & \makecell{AIG Nodes\\(Top1)} & \makecell{AIG Nodes\\(TopN)} & \makecell{AIG Levels\\(Top1)} & \makecell{AIG Levels\\(TopN)} & \makecell{Accuracy\\(Top-1)} & \makecell{Accuracy\\(Top-N)} \\
\midrule
2 &         20 &      No &           95,176 &    95,725.67 (476.25) &               177 &     177.00 (0.00) &   \textbf{49.80} &    \textbf{49.05 (0.65)} \\
4 &         15 &      No &          130,155 &   130,099.33 (201.36) &               202 &     202.00 (0.00) &           48.98 &    48.01 (1.02) \\
4 &         20 &      No &          176,738 &   176,796.33 (190.33) &               213 &     213.00 (0.00) &            48.30 &    48.28 (0.04) \\
3 &         15 &      No &          100,518 &   100,195.67 (284.42) &               184 &     184.00 (0.00) &           47.89 &    47.57 (0.39) \\
3 &         20 &      No &          136,543 &   136,395.33 (128.10) &               195 &     195.00 (0.00) &           47.62 &    47.35 (0.25) \\
2 &         15 &      No &           73,452 &    73,568.00 (180.53) &               166 &     166.00 (0.00) &           45.65 &    45.53 (0.11) \\
3 &         10 &      No &           47,049 &     47,014.00 (30.32) &               164 &     164.00 (0.00) &            44.90 &    44.85 (0.04) \\
\bottomrule
\end{tabular}
\end{adjustbox}
\end{table}

% --------------------------------------------------------------
%\newpage
\begin{table}[ht!]
\centering
\caption{\small Results of the \textbf{random forest logic} on the \textbf{"VACS Model 1"} setup for a quantization scheme of \textbf{32 total bits and 16 fractional bits}.}\label{rf_tb32_vacs_results}
\begin{adjustbox}{width=\textwidth}
\begin{tabular}{ccccccccc}
\toprule
\makecell{Estimators} & \makecell{Max.\\Depth} & \makecell{Bitwise} & \makecell{AIG Nodes\\(Top1)} & \makecell{AIG Nodes\\(TopN)} & \makecell{AIG Levels\\(Top1)} & \makecell{AIG Levels\\(TopN)} & \makecell{Accuracy\\(Top-1)} & \makecell{Accuracy\\(Top-N)} \\
\midrule
\midrule
2 &         10 &      No &           25,4579 &       254,579.00 (0.00) &               354 &     354.00 (0.00) &  \textbf{38.91} &    \textbf{38.91 (0.00)} \\
2 &         15 &     Yes &          1,148,974 &   1,147,517.33 (1862.34) &               350 &     349.33 (0.58) &           37.76 &    35.03 (2.36) \\
4 &         20 &      No &           583,187 &    584,068.33 (1286.80) &               527 &     527.00 (0.00) &           37.07 &    36.39 (0.90) \\
2 &          5 &     Yes &           322,979 &    322,672.33 (1095.67) &               338 &     338.00 (0.00) &           37.01 &    33.88 (3.35) \\
3 &         10 &      No &           280,859 &       280,859.00 (0.00) &               424 &     424.00 (0.00) &           36.46 &    36.46 (0.00) \\
3 &         15 &      No &           389,633 &     390,194.67 (486.44) &               443 &     443.00 (0.00) &           35.99 &    35.01 (0.93) \\
3 &         20 &      No &           504,479 &     504,248.33 (332.66) &               457 &     457.00 (0.00) &           35.85 &    35.72 (0.12) \\
\bottomrule
\end{tabular}
\end{adjustbox}
\end{table}

% --------------------------------------------------------------
%\newpage
\begin{table}[ht!]
\centering
\caption{\small Results of the \textbf{random forest logic} on the \textbf{"VACS Model 2"} setup for a quantization scheme of \textbf{6 total bits and 4 fractional bits}.}\label{rf_tb6_vacs_big_results}
\begin{adjustbox}{width=\textwidth}
\begin{tabular}{ccccccccc}
\toprule
\makecell{Estimators} & \makecell{Max.\\Depth} & \makecell{Bitwise} & \makecell{AIG Nodes\\(Top1)} & \makecell{AIG Nodes\\(TopN)} & \makecell{AIG Levels\\(Top1)} & \makecell{AIG Levels\\(TopN)} & \makecell{Accuracy\\(Top-1)} & \makecell{Accuracy\\(Top-N)} \\
\midrule
3 &          5 &     Yes &           83,697 &     83,982.67 (247.88) &               146 &     146.00 (0.00) &  \textbf{37.69} &    34.17 (3.06) \\
4 &          5 &     Yes &          104,782 &    104,871.00 (105.82) &               158 &     158.00 (0.00) &           35.71 &    33.79 (1.79) \\
2 &         10 &      No &          104,390 &    104,428.00 (194.80) &               153 &     153.00 (0.00) &           35.65 &    34.49 (1.00) \\
4 &         10 &     Yes &         1,043,600 &  1,039,941.33 (3939.50) &               180 &     180.00 (0.00) &           35.31 &    34.08 (1.06) \\
2 &         20 &     Yes &          830,090 &    831,161.00 (931.26) &               194 &     194.00 (0.00) &           35.17 &    34.31 (0.93) \\
3 &         10 &     Yes &          792,146 &    792,785.67 (656.61) &               168 &     168.00 (0.00) &           34.97 &    33.29 (1.46) \\
4 &         10 &      No &          176,146 &    176,089.33 (152.13) &               197 &     197.00 (0.00) &           34.76 &    34.26 (0.45) \\
\bottomrule
\end{tabular}
\end{adjustbox}
\end{table}

% --------------------------------------------------------------
%\newpage
\begin{table}[ht!]
\centering
\caption{\small Results of the \textbf{random forest logic} on the \textbf{"VACS Model 2"} setup for a quantization scheme of \textbf{32 total bits and 16 fractional bits}.}\label{rf_tb32_vacs_big_results}
\begin{adjustbox}{width=\textwidth}
\begin{tabular}{ccccccccc}
\toprule
\makecell{Estimators} & \makecell{Max.\\Depth} & \makecell{Bitwise} & \makecell{AIG Nodes\\(Top1)} & \makecell{AIG Nodes\\(TopN)} & \makecell{AIG Levels\\(Top1)} & \makecell{AIG Levels\\(TopN)} & \makecell{Accuracy\\(Top-1)} & \makecell{Accuracy\\(Top-N)} \\
\midrule
4 &         15 &     Yes &          2,097,694 &  2,095,119.33 (5562.45) &               373 &     373.33 (0.58) &  \textbf{39.39} &    35.26 (3.76) \\
2 &         10 &      No &           254,579 &     254,570.67 (14.43) &               354 &     354.00 (0.00) &           38.91 &    \textbf{38.91 (0.00)} \\
4 &         20 &      No &           584,884 &    584,902.33 (799.66) &               527 &     527.00 (0.00) &           37.07 &    36.51 (0.54) \\
3 &         20 &     Yes &          1,621,931 &  1,625,149.00 (3028.34) &               361 &     361.00 (0.00) &           36.53 &    34.94 (1.88) \\
3 &         10 &      No &           280,859 &      280,859.00 (0.00) &               424 &     424.00 (0.00) &           36.46 &    36.46 (0.00) \\
3 &         10 &     Yes &          1,631,324 &  1,623,684.67 (9490.92) &               361 &     360.67 (0.58) &           36.39 &    34.76 (1.54) \\
3 &         20 &      No &           503,445 &    504,230.67 (698.40) &               457 &     457.00 (0.00) &           36.39 &    35.17 (1.47) \\
\bottomrule
\end{tabular}
\end{adjustbox}
\end{table}

% -------------------------------------------------------------
% \newpage
\clearpage
\subsubsection{LogicNet Evaluation}
The tables provided in the following show the results of the logic creation with LogicNets as intermediate steps. They are based on the real-valued neural network baselines defined in Table \ref{real_nn_baselines_refined} that served as input to the logic pipeline. In each table, the results are sorted in descending manner according to the top-1 model's performance and the column-wise "winner" is marked in bold.

Note that if the number of nodes or levels in the AIG is zero, this means that the LogicNets have learned nothing and its corresponding logic has a constant class prediction.

% --------------------------------------------------------------
%\newpage
\begin{table}[ht!]
\centering
\caption{\small Results of the \textbf{LogicNet logic} on the \textbf{"GIB Model 1"} setup for a quantization scheme of \textbf{7 total bits and 3 fractional bits}.}\label{lgn_gib_small_tb7}
\begin{adjustbox}{width=\textwidth}
\begin{tabular}{ccccccccc}
\toprule
\makecell{Depth} & \makecell{Width} & \makecell{LUT Size} & \makecell{AIG Nodes\\(Top1)} & \makecell{AIG Nodes\\(TopN)} & \makecell{AIG Levels\\(Top1)} & \makecell{AIG Levels\\(TopN)} & \makecell{Accuracy\\(Top-1)} & \makecell{Accuracy\\(Top-N)} \\
\midrule
    4 &  1,000 &        1 &            2,508 &     2,508.00 (0.00) &               106 &     106.00 (0.00) &  \textbf{66.56} &    \textbf{66.56 (0.00)} \\
    2 &   100 &        1 &            2,508 &     2,508.00 (0.00) &               106 &     106.00 (0.00) &           66.56 &    66.56 (0.00) \\
    3 &    50 &        1 &            2,477 &     2,477.00 (0.00) &               110 &     110.00 (0.00) &           65.73 &    65.73 (0.00) \\
    2 &   500 &        1 &            2,477 &     2,477.00 (0.00) &               110 &     110.00 (0.00) &           65.73 &    65.73 (0.00) \\
    3 &  1,000 &        1 &            2,477 &     2,477.00 (0.00) &               110 &     110.00 (0.00) &           65.73 &    65.73 (0.00) \\
    2 &   200 &        2 &            7,033 &  4,688.67 (4,060.50) &               112 &     74.67 (64.66) &           50.17 &    50.11 (0.10) \\
    \midrule
    2 &   500 &        3 &      \textbf{0} &        \textbf{0.00 (0.00)} &        \textbf{0} &       \textbf{0.00 (0.00)} &            50.00 &    50.00 (0.00) \\
\bottomrule
\end{tabular}
\end{adjustbox}
\end{table}

% --------------------------------------------------------------
%\newpage
\begin{table}[ht!]
\centering
\caption{\small Results of the \textbf{LogicNet logic} on the \textbf{"GIB Model 1"} setup for a quantization scheme of \textbf{18 total bits and 10 fractional bits}.}\label{lgn_gib_small_tb18}
\begin{adjustbox}{width=\textwidth}
\begin{tabular}{ccccccccc}
\toprule
\makecell{Depth} & \makecell{Width} & \makecell{LUT Size} & \makecell{AIG Nodes\\(Top1)} & \makecell{AIG Nodes\\(TopN)} & \makecell{AIG Levels\\(Top1)} & \makecell{AIG Levels\\(TopN)} & \makecell{Accuracy\\(Top-1)} & \makecell{Accuracy\\(Top-N)} \\
\midrule
    4 &  1,000 &        1 &           25,536 &      25,536.00 (0.00) &               198 &     198.00 (0.00) &  \textbf{65.56} &     \textbf{65.56 (0.00)} \\
    2 &   100 &        1 &           25,536 &      25,536.00 (0.00) &               198 &     198.00 (0.00) &           65.56 &    65.56 (0.00) \\
    2 &   500 &        1 &           21,810 &      21,810.00 (0.00) &               192 &     192.00 (0.00) &            64.40 &    64.40 (0.00) \\
    3 &  1,000 &        1 &           21,810 &      21,810.00 (0.00) &               192 &     192.00 (0.00) &            64.40 &    64.40 (0.00) \\
    3 &    50 &        1 &           21,810 &      21,810.00 (0.00) &               192 &     192.00 (0.00) &            64.40 &    64.40 (0.00) \\
    2 &    50 &        2 &           62,998 &     62,980.00 (41.07) &               199 &     196.33 (6.43) &            50.50 &    50.33 (0.29) \\
    2 &   500 &        2 &           61,787 &  41,219.33 (35,697.01) &               189 &   131.33 (114.02) &           50.17 &    50.06 (0.10) \\
\bottomrule
\end{tabular}
\end{adjustbox}
\end{table}

% --------------------------------------------------------------
%\newpage
\begin{table}[ht!]
\centering
\caption{\small Results of the \textbf{LogicNet logic} on the \textbf{"GIB Model 2"} setup for a quantization scheme of \textbf{6 total bits and 4 fractional bits}.}\label{lgn_gib_big_tb6}
\begin{adjustbox}{width=\textwidth}
\begin{tabular}{ccccccccc}
\toprule
\makecell{Depth} & \makecell{Width} & \makecell{LUT Size} & \makecell{AIG Nodes\\(Top1)} & \makecell{AIG Nodes\\(TopN)} & \makecell{AIG Levels\\(Top1)} & \makecell{AIG Levels\\(TopN)} & \makecell{Accuracy\\(Top-1)} & \makecell{Accuracy\\(Top-N)} \\
\midrule
    3 &    50 &        2 &           13,501 &   13,466.00 (45.92) &               131 &     129.67 (2.31) &   \textbf{64.40} &    \textbf{54.80 (8.31)} \\
    3 &   500 &        3 &           25,292 &   25,279.00 (28.79) &               139 &     138.67 (0.58) &           51.82 &    50.61 (1.05) \\
    2 &   500 &        2 &            9,815 &     98,10.00 (9.54) &               124 &     123.00 (1.00) &           51.49 &    49.12 (2.92) \\
    2 &    50 &        2 &            9,727 &    9,715.67 (16.29) &               133 &     132.33 (1.15) &           50.99 &    50.27 (0.97) \\
    2 &   100 &        2 &            8,157 &    8,152.33 (12.66) &               125 &     124.33 (1.15) &           50.99 &    50.05 (0.82) \\
    2 &    50 &        3 &           18,491 &   18,478.33 (16.26) &               137 &     137.00 (0.00) &           50.33 &    50.11 (0.19) \\
    4 &    50 &        3 &           25,322 &  25,176.00 (130.43) &               137 &     135.00 (2.00) &           50.33 &    50.11 (0.19) \\
\bottomrule
\end{tabular}
\end{adjustbox}
\end{table}

% --------------------------------------------------------------
%\newpage
\begin{table}[ht!]
\centering
\caption{\small Results of the \textbf{LogicNet logic} on the \textbf{"GIB Model 2"} setup for a quantization scheme of \textbf{12 total bits and 8 fractional bits}.}\label{lgn_gib_big_tb12}
\begin{adjustbox}{width=\textwidth}
\begin{tabular}{ccccccccc}
\toprule
\makecell{Depth} & \makecell{Width} & \makecell{LUT Size} & \makecell{AIG Nodes\\(Top1)} & \makecell{AIG Nodes\\(TopN)} & \makecell{AIG Levels\\(Top1)} & \makecell{AIG Levels\\(TopN)} & \makecell{Accuracy\\(Top-1)} & \makecell{Accuracy\\(Top-N)} \\
\midrule
    2 &   500 &        2 &           41,867 &  13,955.67 (24,171.92) &               165 &     55.00 (95.26) &   \textbf{50.50} &    \textbf{50.17 (0.29)} \\
    3 &    50 &        3 &           93,893 &     93,933.00 (75.43) &               187 &     187.00 (2.00) &           50.17 &    50.06 (0.10) \\
    3 &   200 &        2 &           50,187 &     50,233.33 (67.68) &               175 &     178.00 (5.20) &           50.17 &    50.00 (0.17) \\
    \midrule
    2 &    50 &        1 &      \textbf{0} &          \textbf{0.00 (0.00)} &        \textbf{0} &       \textbf{0.00 (0.00)} &            50.00 &    50.00 (0.00) \\
    4 &   100 &        2 &           87,276 &     87,261.67 (55.90) &               177 &     176.00 (3.61) &            50.00 &    50.00 (0.00) \\
    3 &    50 &        1 &               0 &          0.00 (0.00) &                 0 &       0.00 (0.00) &            50.00 &    50.00 (0.00) \\
    3 &    50 &        2 &           56,822 &  18,940.67 (32,806.20) &               173 &     57.67 (99.88) &            50.00 &    50.00 (0.00) \\
\bottomrule
\end{tabular}
\end{adjustbox}
\end{table}

% --------------------------------------------------------------
%\newpage
\begin{table}[ht!]
\centering
\caption{\small Results of the \textbf{LogicNet logic} on the \textbf{"VACS Model 1"} setup for a quantization scheme of \textbf{7 total bits and 3 fractional bits}.}\label{lgn_vacs_small_tb7}
\begin{adjustbox}{width=\textwidth}
\begin{tabular}{ccccccccc}
\toprule
\makecell{Depth} & \makecell{Width} & \makecell{LUT Size} & \makecell{AIG Nodes\\(Top1)} & \makecell{AIG Nodes\\(TopN)} & \makecell{AIG Levels\\(Top1)} & \makecell{AIG Levels\\(TopN)} & \makecell{Accuracy\\(Top-1)} & \makecell{Accuracy\\(Top-N)} \\
\midrule
    2 &   500 &        3 &            7,473 &    7,445.33 (65.07) &                89 &      84.00 (8.66) &   \textbf{33.40} &    \textbf{33.35 (0.04)} \\
    2 &  1,000 &        3 &            7,397 &    7,378.33 (37.65) &                78 &      74.00 (6.08) &            33.40 &    33.35 (0.04) \\
    \midrule
    2 &    50 &        3 &      \textbf{0} &  4,738.67 (4103.81) &        \textbf{0} &     \textbf{46.67 (40.41)} &           33.33 &    33.33 (0.00) \\
    3 &    50 &        5 &           10,038 &    9,963.33 (77.66) &                75 &      75.33 (0.58) &           33.33 &    33.33 (0.00) \\
    2 &  1,000 &        4 &            8,476 &    8,458.00 (25.36) &                85 &      83.33 (1.53) &           33.33 &    33.33 (0.00) \\
    4 &   100 &        4 &            9,895 &    9,970.67 (70.68) &                74 &      76.33 (2.08) &           33.33 &    33.33 (0.00) \\
    2 &  1,000 &        5 &               0 &  5,634.33 (4,879.48) &                 0 &     47.67 (41.28) &           33.33 &    33.33 (0.00) \\
\bottomrule
\end{tabular}
\end{adjustbox}
\end{table}

% --------------------------------------------------------------
%\newpage
\begin{table}[ht!]
\centering
\caption{\small Results of the \textbf{LogicNet logic} on the \textbf{"VACS Model 1"} setup for a quantization scheme of \textbf{32 total bits and 16 fractional bits}.}\label{lgn_vacs_small_tb32}
\begin{adjustbox}{width=\textwidth}
\begin{tabular}{ccccccccc}
\toprule
\makecell{Depth} & \makecell{Width} & \makecell{LUT Size} & \makecell{AIG Nodes\\(Top1)} & \makecell{AIG Nodes\\(TopN)} & \makecell{AIG Levels\\(Top1)} & \makecell{AIG Levels\\(TopN)} & \makecell{Accuracy\\(Top-1)} & \makecell{Accuracy\\(Top-N)} \\
\midrule
    3 &  1,000 &        3 &          186,572 &    186,531.00 (65.89) &               274 &     270.67 (3.51) &   \textbf{33.40} &     \textbf{33.35 (0.04)} \\
    2 &   500 &        3 &          136,429 &    136,462.67 (33.02) &               260 &     262.33 (2.52) &            33.40 &    33.31 (0.10) \\
    4 &   200 &        3 &          186,985 &    186,999.00 (37.99) &               277 &     277.00 (1.00) &            33.40 &    33.35 (0.04) \\
    \midrule
    3 &   200 &        2 &      \textbf{0} &          \textbf{0.00 (0.00)} &        \textbf{0} &        \textbf{0.00 (0.00)} &           33.33 &    33.33 (0.00) \\
    4 &   100 &        3 &          186,900 &    186,941.00 (56.15) &               278 &     276.33 (3.79) &           33.33 &    33.33 (0.00) \\
    4 &    50 &        3 &          186,363 &    186,426.33 (57.42) &               237 &     239.67 (3.06) &           33.33 &    33.33 (0.00) \\
    3 &    50 &        2 &          135,310 &     135,298.67 (9.87) &               244 &     243.00 (2.65) &           33.33 &    33.33 (0.00) \\
\bottomrule
\end{tabular}
\end{adjustbox}
\end{table}

% --------------------------------------------------------------
%\newpage
\begin{table}[ht!]
\centering
\caption{\small Results of the \textbf{LogicNet logic} on the \textbf{"VACS Model 2"} setup for a quantization scheme of \textbf{6 total bits and 4 fractional bits}.}\label{lgn_vacs_big_tb6}
\begin{adjustbox}{width=\textwidth}
\begin{tabular}{ccccccccc}
\toprule
\makecell{Depth} & \makecell{Width} & \makecell{LUT Size} & \makecell{AIG Nodes\\(Top1)} & \makecell{AIG Nodes\\(TopN)} & \makecell{AIG Levels\\(Top1)} & \makecell{AIG Levels\\(TopN)} & \makecell{Accuracy\\(Top-1)} & \makecell{Accuracy\\(Top-N)} \\
\midrule
    3 &   200 &        2 &      \textbf{0} &       \textbf{0.00 (0.00)} &        \textbf{0} &        \textbf{0.00 (0.00)} &  \textbf{33.33} &     \textbf{33.33 (0.00)} \\
    2 &  1,000 &        1 &               0 &        0.00 (0.00) &                 0 &       0.00 (0.00) &           33.33 &    33.33 (0.00) \\
    2 &  1,000 &        2 &               0 &        0.00 (0.00) &                 0 &       0.00 (0.00) &           33.33 &    33.33 (0.00) \\
    4 &   100 &        2 &           12,429 &   12,480.33 (99.49) &                82 &      83.00 (8.54) &           33.33 &    33.33 (0.00) \\
    2 &  1,000 &        3 &               0 &  2,620.00 (4,537.97) &                 0 &     29.67 (51.38) &           33.33 &    33.33 (0.00) \\
    4 &   100 &        3 &           26,802 &   26,802.33 (18.50) &               101 &     103.00 (1.73) &           33.33 &    33.33 (0.00) \\
    3 &    50 &        1 &               0 &        0.00 (0.00) &                 0 &       0.00 (0.00) &           33.33 &    33.33 (0.00) \\
\bottomrule
\end{tabular}
\end{adjustbox}
\end{table}

% --------------------------------------------------------------
% \newpage
\begin{table}[ht!]
\centering
\caption{\small Results of the \textbf{LogicNet logic} on the \textbf{"VACS Model 2"} setup for a quantization scheme of \textbf{32 total bits and 16 fractional bits}.}\label{lgn_vacs_big_tb32}
\begin{adjustbox}{width=\textwidth}
\begin{tabular}{ccccccccc}
\toprule
\makecell{Depth} & \makecell{Width} & \makecell{LUT Size} & \makecell{AIG Nodes\\(Top1)} & \makecell{AIG Nodes\\(TopN)} & \makecell{AIG Levels\\(Top1)} & \makecell{AIG Levels\\(TopN)} & \makecell{Accuracy\\(Top-1)} & \makecell{Accuracy\\(Top-N)} \\
\midrule
    3 &   500 &        3 &          389,577 &      389,558.33 (19.01) &               264 &     264.33 (1.53) &  \textbf{33.47} &    \textbf{33.38 (0.08)} \\
    \midrule
    2 &    50 &        1 &      \textbf{0} &            \textbf{0.00 (0.00)} &        \textbf{0} &       \textbf{0.00 (0.00)} &           33.33 &    33.33 (0.00) \\
    3 &   100 &        2 &          126,671 &      126,674.33 (13.32) &               248 &     249.00 (1.73) &           33.33 &    33.33 (0.00) \\
    2 &  1,000 &        3 &          166,219 &      166,226.33 (48.42) &               251 &     254.33 (7.57) &           33.33 &    33.33 (0.00) \\
    3 &    50 &        1 &               0 &            0.00 (0.00) &                 0 &       0.00 (0.00) &           33.33 &    33.33 (0.00) \\
    4 &   100 &        1 &               0 &            0.00 (0.00) &                 0 &       0.00 (0.00) &           33.33 &    33.33 (0.00) \\
    3 &    50 &        2 &          141,596 &     141,990.33 (341.51) &               187 &    233.00 (39.84) &           33.33 &    33.33 (0.00) \\
\bottomrule
\end{tabular}
\end{adjustbox}
\end{table}

% -------------------------------------------------------
\clearpage
\subsection{Interpretable Report}
\label{interpretable_report}

\begin{figure*}[ht]
\begin{center}
\centerline{\includegraphics[width=\textwidth]{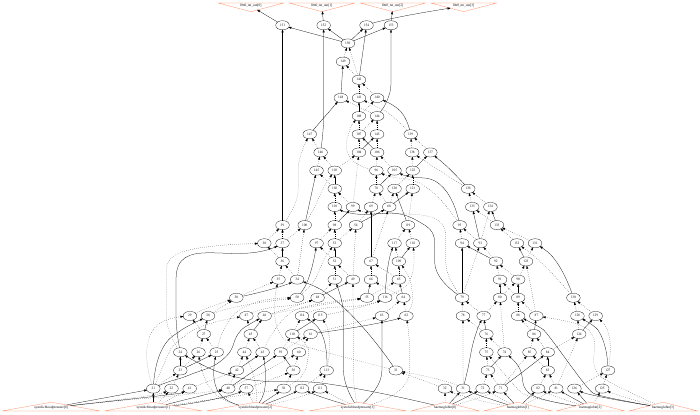}}
\caption{Example of the AIG that is derived from a simple logic module. The visualization was exported after creating the AIG with ABC \cite{abc}.}
\label{l0n0_figure}
\end{center}
\end{figure*}

In the following, an example of how interpretable reports from the logic can look like is reported. It is derived for a single node $l0;n0$ of the first layer from a direct translation of a neural network to arithmetic logic. It receives four input features of the GIB data set: the systolic blood pressure, the haemoglobin level, the urea level and the creatinine level. The quantization scheme has 4 total bits and 2 fractional bits. The Boolean equations derived from the AIG of the logic module can be found in the following. The derived AIG has 134 nodes and can be visualized as shown in Figure \ref{l0n0_figure}. From the report it can seen that in this example there is no dependency on the variables creatinine and urea, which means that they are internally treated as don't-cares. Similar reports can be created from any arbitrary concatenation of logic modules, allowing simulations and interpretations also of only intermediate layers or nodes or measuring variables' influences.

We, however, note that the report is human readable but usually not derived for this purpose, except for use-cases where the logic is small enough such that a system verification process can be executed manually. The example should mainly provide insights on how the AIG nodes' formulas look like in terms of factored AND and OR gates. Running SAT-solvers on such logic functions to find satisfying assignments can serve the process of system verification. Also from this small example it can already be seen that missing dependencies on features can be found which can help on curating data sets.

\vspace{5mm}
\begin{tcolorbox}[%
    standard jigsaw,
    opacityback=0,
    enhanced, 
    breakable,
    frame hidden,
    overlay broken = {
        \draw[line width=0.2mm, black, rounded corners]
        (frame.north west) rectangle (frame.south east);},
    ]{}
    
\textbf{Logic Report:} l0n0{\_}nn \\
\noindent\rule{\columnwidth}{0.4pt}
\newline

\textbf{Inputs:}\\
\noindent\rule{\columnwidth}{0.4pt}
Input 0:	systolic\\
Input 1:	haemoglobin\\
Input 2:	urea\\
Input 3:	creatinine

\textbf{Outputs:}\\
\noindent\rule{\columnwidth}{0.4pt}
Output:	l0n0{\_}nn{\_}out\\

\textbf{Equations:}\\
\noindent\rule{\columnwidth}{0.4pt}
n21 = systolic[0] AND systolic[1];\\
n22 = NOT systolic[0] AND NOT systolic[1];\\
n23 = NOT n21 AND NOT n22;\\
n24 = haemoglobin[0] AND n23;\\
n25 = systolic[2] AND n23;\\
n26 = NOT systolic[2] AND NOT n23;\\
n27 = NOT n25 AND NOT n26;\\
n28 = n21 AND n27;\\
n29 = NOT n21 AND NOT n27;\\
n30 = NOT n28 AND NOT n29;\\
n31 = haemoglobin[0] AND haemoglobin[1];\\
n32 = NOT haemoglobin[0] AND NOT haemoglobin[1];\\
n33 = NOT n31 AND NOT n32;\\
n34 = n30 AND n33;\\
n35 = NOT n30 AND NOT n33;\\
n36 = NOT n34 AND NOT n35;\\
n37 = n24 AND n36;\\
n38 = NOT n24 AND NOT n36;\\
n39 = NOT n37 AND NOT n38;\\
n40 = NOT systolic[0] AND systolic[1];\\
n41 = systolic[0] AND NOT systolic[1];\\
n42 = NOT n40 AND NOT n41;\\
n43 = systolic[2] AND n42;\\
n44 = NOT systolic[2] AND NOT n42;\\
n45 = NOT n43 AND NOT n44;\\
n46 = n21 AND n45;\\
n47 = NOT n21 AND NOT n45;\\
n48 = NOT n46 AND NOT n47;\\
n49 = NOT systolic[3] AND n48;\\
n50 = NOT n25 AND NOT n28;\\
n51 = systolic[3] AND NOT n48;\\
n52 = NOT n49 AND NOT n51;\\
n53 = NOT n50 AND n52;\\
n54 = NOT n49 AND NOT n53;\\
n55 = NOT n43 AND NOT n46;\\
n56 = NOT systolic[1] AND systolic[2];\\
n57 = systolic[1] AND NOT systolic[2];\\
n58 = NOT n56 AND NOT n57;\\
n59 = n40 AND n58;\\
n60 = NOT n40 AND NOT n58;\\
n61 = NOT n59 AND NOT n60;\\
n62 = NOT systolic[3] AND n61;\\
n63 = systolic[3] AND NOT n61;\\
n64 = NOT n62 AND NOT n63;\\
n65 = NOT n55 AND n64;\\
n66 = n55 AND NOT n64;\\
n67 = NOT n65 AND NOT n66;\\
n68 = n54 AND NOT n67;\\
n69 = NOT n54 AND n67;\\
n70 = NOT n68 AND NOT n69;\\
n71 = NOT haemoglobin[0] AND haemoglobin[1];\\
n72 = haemoglobin[0] AND NOT haemoglobin[1];\\
n73 = NOT n71 AND NOT n72;\\
n74 = haemoglobin[2] AND n73;\\
n75 = NOT haemoglobin[2] AND NOT n73;\\
n76 = NOT n74 AND NOT n75;\\
n77 = n31 AND n76;\\
n78 = NOT n31 AND NOT n76;\\
n79 = NOT n77 AND NOT n78;\\
n80 = NOT n74 AND NOT n77;\\
n81 = NOT haemoglobin[1] AND haemoglobin[2];\\
n82 = haemoglobin[1] AND NOT haemoglobin[2];\\
n83 = NOT n81 AND NOT n82;\\
n84 = n71 AND n83;\\
n85 = NOT n71 AND NOT n83;\\
n86 = NOT n84 AND NOT n85;\\
n87 = NOT haemoglobin[3] AND n86;\\
n88 = haemoglobin[3] AND NOT n86;\\
n89 = NOT n87 AND NOT n88;\\
n90 = NOT n80 AND n89;\\
n91 = n80 AND NOT n89;\\
n92 = NOT n90 AND NOT n91;\\
n93 = NOT n79 AND NOT n92;\\
n94 = n79 AND n92;\\
n95 = NOT n93 AND NOT n94;\\
n96 = NOT n70 AND NOT n95;\\
n97 = n50 AND NOT n52;\\
n98 = NOT n53 AND NOT n97;\\
n99 = NOT n79 AND n98;\\
n100 = NOT n34 AND NOT n37;\\
n101 = n79 AND NOT n98;\\
n102 = NOT n99 AND NOT n101;\\
n103 = NOT n100 AND n102;\\
n104 = NOT n99 AND NOT n103;\\
n105 = n70 AND n95;\\
n106 = NOT n96 AND NOT n105;\\
n107 = NOT n104 AND n106;\\
n108 = NOT n96 AND NOT n107;\\
n109 = NOT n62 AND NOT n65;\\
n110 = NOT n56 AND NOT n59;\\
n111 = NOT systolic[2] AND NOT systolic[3];\\
n112 = systolic[2] AND systolic[3];\\
n113 = NOT n111 AND NOT n112;\\
n114 = NOT n110 AND n113;\\
n115 = n110 AND NOT n113;\\
n116 = NOT n114 AND NOT n115;\\
n117 = NOT n109 AND n116;\\
n118 = n109 AND NOT n116;\\
n119 = NOT n117 AND NOT n118;\\
n120 = NOT n68 AND n119;\\
n121 = n68 AND NOT n119;\\
n122 = NOT n120 AND NOT n121;\\
n123 = NOT n87 AND NOT n90;\\
n124 = NOT n81 AND NOT n84;\\
n125 = NOT haemoglobin[2] AND NOT haemoglobin[3];\\
n126 = haemoglobin[2] AND haemoglobin[3];\\
n127 = NOT n125 AND NOT n126;\\
n128 = NOT n124 AND n127;\\
n129 = n124 AND NOT n127;\\
n130 = NOT n128 AND NOT n129;\\
n131 = NOT n123 AND n130;\\
n132 = n123 AND NOT n130;\\
n133 = NOT n131 AND NOT n132;\\
n134 = NOT n93 AND n133;\\
n135 = n93 AND NOT n133;\\
n136 = NOT n134 AND NOT n135;\\
n137 = n122 AND n136;\\
n138 = NOT n122 AND NOT n136;\\
n139 = NOT n137 AND NOT n138;\\
n140 = NOT n108 AND n139;\\
n141 = n108 AND NOT n139;\\
n142 = NOT n140 AND NOT n141;\\
n143 = n104 AND NOT n106;\\
n144 = NOT n107 AND NOT n143;\\
n145 = n100 AND NOT n102;\\
n146 = NOT n103 AND NOT n145;\\
n147 = NOT n39 AND NOT n146;\\
n148 = NOT n144 AND n147;\\
n149 = NOT n142 AND n148;\\
n150 = NOT n142 AND NOT n149;\\
l0n0{\_}nn{\_}out[0] = n39 AND n150;\\
l0n0{\_}nn{\_}out[1] = n146 AND n150;\\
l0n0{\_}nn{\_}out[2] = n144 AND n150;\\
l0n0{\_}nn{\_}out[3] = n142 AND n150;\\

\textbf{No Dependency On:}\\
\noindent\rule{\columnwidth}{0.4pt}
Input 2:	urea\\
Input 3:	creatinine
\end{tcolorbox}

\vspace{5mm}

% -------------------------------------------------------

As a proof of concept, Table \ref{minisat_result} provides the computing times for MiniSAT \cite{een2003extensible} running on AIGs of different sizes, while searching for satisfying assignments of the underlying logic function. The runtimes are in the magnitude of minutes to a few hours.

\begin{table}[ht]
\caption{\small Results of the computing times for MiniSAT \cite{een2003extensible} to find satisfying arguments for multiple settings of bloated and reduced logic - ordered by the number of AIG nodes.}
\label{minisat_result}
\begin{center}
\begin{adjustbox}{width=\textwidth}
\begin{tabular}{cccccc}
\toprule
Logic Translation \\($m$ total bits, $i$ fractional bits) & Settings & AIG Nodes & AIG Levels & MiniSAT time \\
\midrule
Arithmetic Circuit ($32$, $16$) & - & 2,516,251 & 1,113 & 18,813.05 s (313.55 min) \\ % 01
Random Forest ($8$, $6$) & Estimators: 4; Max. Depth: 10 & 912,589 & 275 & 22,310.16 s (371.84 min) \\ % 03
LogicNet ($8$, $6$) & Depth: 4; Width: 100; LUT-Size: 8 & 256,910 & 212 & 1,264.43 s (21.07 min) \\ %AIG Opt. ! % 02
Arithmetic Circuit ($8$, $4$) & - & 210,457 & 477 & 76.08 s (1.27 min) \\ % 06
Random Forest ($8$, $6$) & Estimators: 2; Max. Depth: 5 & 198,378 & 227 & 5,226.24 s (87.10 min) \\ % 05
LogicNet ($8$, $6$) & Depth: 2; Width: 200; LUT-Size: 4 & 56,738 & 184 & 4,880.07 s (81.33 min) \\ %AIG Opt. ! % 04
\bottomrule
\end{tabular}
\end{adjustbox}
\end{center}
\end{table}

\end{document}